\documentclass[pdflatex,sn-mathphys-num]{sn-jnl}

\usepackage{graphicx}%
\usepackage{multirow}%
\usepackage{amsmath,amssymb,amsfonts}%
\usepackage{amsthm}%
\usepackage{mathrsfs}%
\usepackage[title]{appendix}%
\usepackage{xcolor}%
\usepackage{textcomp}%
\usepackage{manyfoot}%
\usepackage{booktabs}%
\usepackage{algorithm}%
\usepackage{algorithmicx}%
\usepackage{algpseudocode}%
\usepackage{listings}%
\usepackage{enumitem}
\usepackage{multicol}
\usepackage{url}
\usepackage{soul}
\usepackage{comment}
\usepackage{makecell}
\usepackage{tablefootnote}

\theoremstyle{thmstyleone}%
\theoremstyle{thmstyletwo}%

\theoremstyle{thmstylethree}%

\begin{document}

\title[Toward Fairness in Machine Learning Models for Predicting Treatment Retention and Premature Discontinuation in Medication for Opioid Use Disorder]{Toward Fairness in Machine Learning Models for Predicting Treatment Retention and Premature Discontinuation in Medication for Opioid Use Disorder}


\author*[1]{\fnm{Tongnian} \sur{Wang}}\email{tongnian-wang@utc.edu}

\author[3]{\fnm{Carolina} \sur{Vivas-Valencia}}\email{carolina.vivasvalencia@utsa.edu}

\author[4,5]{\fnm{Cici} \sur{Bauer}}\email{cici.x.bauer@uth.tmc.edu}

\author[6]{\fnm{Yanmin} \sur{Gong}}\email{yanmin.gong@tamu.edu}

\author[2]{\fnm{Kim-Kwang Raymond} \sur{Choo}}\email{raymond.choo@utsa.edu}

\author[2]{\fnm{Yuanxiong} \sur{Guo}}\email{yuanxiong.guo@utsa.edu}

\affil*[1]{\orgdiv{Gary W. Rollins College of Business}, \orgname{The University of Tennessee at Chattanooga}, \orgaddress{\city{Chattanooga}, \state{Tennessee}, \country{USA}}}

\affil[2]{\orgdiv{Department of Information Systems and Cybersecurity}, \orgname{The University of Texas at San Antonio}, \orgaddress{\city{San Antonio}, \state{Texas}, \country{USA}}}

\affil[3]{\orgdiv{Department of Biomedical Engineering and Chemical Engineering}, \orgname{The University of Texas at San Antonio}, \orgaddress{\city{San Antonio}, \state{Texas}, \country{USA}}}

\affil[4]{\orgdiv{Department of Biostatistics and Data Science, School of Public Health}, \orgname{The University of Texas Health Science Center at Houston}, \orgaddress{\city{Houston}, \state{Texas}, \country{USA}}}

\affil[5]{\orgdiv{Center for Spatial-Temporal Modeling for Applications in Population Sciences, School of Public Health}, \orgname{The University of Texas Health Science Center at Houston}, \orgaddress{\city{Houston}, \state{Texas}, \country{USA}}}

\affil[6]{\orgdiv{School of Engineering Medicine}, \orgname{Texas A\&M University}, \orgaddress{\city{Houston}, \state{Texas}, \country{USA}}}




\abstract{
Persistent low retention and completion rates in medications for opioid use disorder (MOUD) have driven the use of machine learning (ML) models to predict retention and identify patients at risk of premature discontinuation. However, the fairness of these models across patient populations remains largely unexplored, raising concerns about their application in treatment decision support. This study systematically assesses algorithmic fairness in ML models for predicting MOUD retention and premature discontinuation and investigates the effectiveness of bias mitigation techniques. 
Using the cross-sectional Treatment Episode Data Set–Discharges (TEDS-D), which includes treatment episodes for individuals in the U.S. discharged between 2015 and 2019, we trained four ML models to predict premature treatment discontinuation and retention beyond 180 days among individuals receiving outpatient MOUD. We evaluated overall performance and subgroup-level error rates across patient subgroups defined by race, ethnicity, age, and sex, complemented by model explanation analyses. We further assessed pre-processing, in-processing, and post-processing bias mitigation techniques and their effects on both fairness and predictive performance. 
The models exhibited substantial performance differences across patient subgroups, including overestimation of the likelihood of premature discontinuation for Black patients and of treatment retention beyond 180 days for older patients. Model explanation analyses further identified race and age as influential predictors, but their impacts on model predictions varied substantially across patient subgroups. Bias mitigation strategies reduced specific fairness gaps but often introduced trade-offs, such as increased error rates for other subgroups or reductions in overall predictive performance. 
These findings demonstrate that ML models for MOUD outcome prediction can exhibit subgroup-level performance gaps even when overall predictive performance appears acceptable and that bias mitigation can reduce, but not fully eliminate, these gaps without trade-offs. By demonstrating the importance of fairness-aware evaluation and transparent reporting of subgroup performance, this study provides practical insights for the responsible and context-sensitive use of ML models for risk stratification and care prioritization in MOUD treatment settings. }



\maketitle

\section{Introduction}
Opioid use disorder (OUD) continues to represent a critical public health crisis in the United States, with far-reaching consequences for population health, healthcare systems, and the broader economy \cite{volkow2021changing}. OUD is a chronic, relapsing condition marked by persistent opioid use, including both prescribed and illicit opioids, despite adverse physical, psychological, and social consequences \cite{florence2021economic}. Recent estimates indicate that approximately 6 million individuals in the United States were affected by OUD in 2022\cite{moudstudy2024}. In 2023 alone, opioid-involved overdoses accounted for more than 81,000 deaths, representing roughly 75\% of all drug overdose fatalities nationwide \cite{pdodc2025}. Although data from 2024 suggest a modest decline in opioid-related overdose deaths, indicators of ongoing harm persist \cite{usoud2025}. 
In particular, emergency department visits for non-fatal opioid overdoses have continued to increase, rising by approximately 8\% in February 2025 \cite{DOSESYS}. 

Amid this public health emergency, medications for OUD (MOUD)—including methadone, buprenorphine, and naltrexone—are among the most effective interventions. These medication treatments have been shown to significantly reduce both overdose-related and all-cause mortality while supporting long-term stabilization and recovery \cite{mccarty2014substance}. 
Yet, their effectiveness depends heavily on patient engagement and sustained treatment retention. 
Despite their proven efficacy, many patients discontinue therapy prematurely and struggle to remain engaged in MOUD, making long-term retention a persistent and complex challenge, particularly in outpatient treatment settings where sustained engagement is essential for achieving effective outcomes \cite{morgan2018injectable,baird2023determinants}. Short-term regimens or early tapers are often associated with increased risks of relapse, overdose, and reduced long-term stability \cite{timko2016retention}. 
Prior research has shown that treatment discontinuation reflects a complex interplay of clinical, behavioral, and structural factors, including co-occurring mental health conditions, unstable housing or transportation, employment constraints, stigma, limited access to supportive services, as well as demographic characteristics such as race and age \cite{dong2023racial,samples2018risk}. 
These multifaceted factors make it difficult to identify which individuals are most likely to remain engaged in and successfully complete outpatient MOUD treatment \cite{baird2023determinants}. 
As a result, evidence-based interventions known to improve retention, such as proactive appointment reminders and check-ins, early engagement with counseling or behavioral therapy, transportation support or telehealth adjustments, more frequent dosing supervision for clinically unstable patients, and peer recovery coaching or contingency-management interventions, often cannot be delivered uniformly across patients \cite{jacobson2020organizational}. This challenge is further compounded by the significant resource constraints under which many MOUD programs operate, limiting their capacity to provide intensive follow-up and support to all patients \cite{dong2023racial,jacobson2020organizational}. 

In this context, accurate prediction of patients at high risk of premature discontinuation can play a critical role in guiding clinical prioritization and resource allocation. 
Machine learning (ML) models can leverage large-scale datasets to identify patterns and risk factors associated with treatment disengagement \cite{hasan2021machine,lopez2024predicting}. 
By enabling early risk identification, ML models can support several practical uses in MOUD treatment settings, including risk stratification at intake to inform initial treatment planning, flagging early signs of disengagement within the first days or weeks of treatment, and triggering clinical decision support for targeted follow-up interventions \cite{stafford2022identifying,warren2022using}. 

However, differences in MOUD access and retention across patient sociodemographic subgroups are often reflected in the data used to train these models. 
In addition to individual-level complexity, prior research has identified consistent patterns in treatment retention and completion rates across different patient population subgroups. 
Studies have shown that certain patient sociodemographic subgroups may experience lower rates of MOUD initiation and a higher likelihood of premature discontinuation \cite{stahler2021racial,krawczyk2021stays,mintz2020age}. These patterns are often influenced by factors such as limited provider availability, lack of insurance coverage, and logistical obstacles related to transportation, scheduling, or clinic accessibility \cite{james2018opioid,acevedo2020barriers,pinedo2019current,stahler2021racial}. 
Additional external pressures, such as financial hardship or limited access to nearby treatment programs, can further reduce the likelihood of completing a full course of care \cite{james2018opioid,abraham2018geographic}. 
As a result, there is growing concern that ML models may unintentionally produce inconsistent or less reliable predictions across different patient subgroups, leading to variation in model performance that could affect clinical decision-making—an important consideration known as algorithmic fairness \cite{obermeyer2019dissecting,paulus2020predictably,chen2023algorithmic}. 
These inconsistencies can arise at any stage of the ML model development cycle, ranging from data collection to model training and evaluation, and may result in treatment recommendations that do not generalize well across all patient subgroups \cite{obermeyer2019dissecting,thompson2021bias}. 
When ML models consistently misestimate treatment outcomes for certain patient subgroups, patients may receive care that is not well-aligned with their clinical profiles or treatment needs, ultimately limiting the overall reliability and clinical utility of these models. 

Although ML models have been extensively studied in MOUD research, prior work has paid limited attention to treatment outcome prediction specifically in outpatient treatment settings, as well as to systematic evaluation of how these models perform across different patient subgroups \cite{stafford2022identifying,warren2022using,dong2021identifying,acion2017use,lo2019evaluation,guo2025state}. 
While the broader healthcare ML literature has increasingly explored issues related to algorithmic fairness and reliability across patient subgroups \cite{gianfrancesco2018potential,juhn2022assessing,wang2024analyzing}, there remains a significant gap in understanding how such concerns apply to ML models specifically designed for MOUD retention and completion. 
Furthermore, although numerous strategies have been developed to improve model fairness across subgroups \cite{huang2022evaluation,mehrabi2021survey}, their effectiveness in the context of MOUD retention and premature discontinuation prediction remain largely unexplored. 
Addressing these gaps is crucial to developing reliable and clinically useful ML decision support tools in addiction treatment. 
Therefore, this study seeks to address the following research questions: 
\begin{itemize}
\item How accurately can ML models predict treatment retention and premature discontinuation in outpatient MOUD treatment settings using routinely collected administrative data? 
\item To what extent do ML models show performance inconsistency across different patient subgroups when predicting MOUD retention and premature discontinuation?
\item How do model predictions and underlying feature attributions differ across patient subgroups, and what do these differences reveal about patterns in model behavior?
\item How effective are bias mitigation techniques—such as pre-processing reweighting, in-processing constraint optimization, and post-processing threshold adjustment—in improving model fairness across patient subgroups while maintaining overall predictive accuracy? 
\end{itemize}

To answer these questions, this study systematically assesses the fairness and reliability of ML models developed to predict MOUD retention and premature discontinuation. 
We evaluate the predictive performance and fairness of ML models across patient subgroups defined by race, ethnicity, age, and sex, investigate subgroup-level differences in model behavior using model explanation techniques, and evaluate representative bias mitigation strategies spanning the pre-processing, in-processing, and post-processing stages of the ML pipeline. 
Through this comprehensive analysis, our goal is to improve understanding of how ML models perform across different patient populations in the context of MOUD, and to offer practical guidance for developing more consistent and clinically reliable predictive tools to support treatment planning and decision-making. 
In summary, this paper makes the following key contributions: 
\begin{enumerate}
\item We develop and evaluate ML models to predict MOUD treatment outcomes in outpatient treatment settings, including treatment retention and premature discontinuation. This fills an important gap in the MOUD prediction literature, which has largely focused on mixed treatment settings. Outpatient MOUD care is less structured, relies more heavily on sustained patient self-engagement, and is more strongly influenced by dynamic social and contextual factors, making outcome prediction both more challenging and more operationally consequential than in other treatment settings. 

\item We conduct a comprehensive empirical evaluation of algorithmic fairness in ML models predicting MOUD treatment retention and premature discontinuation. By integrating subgroup-level performance assessment with model explanation analyses, we identify systematic differences in error rates, feature importance, and predictive behavior across patient sociodemographic subgroups. This analysis contributes empirical evidence regarding the fairness and reliability of ML models for MOUD outcome prediction. 

\item We assess the effectiveness of widely used bias mitigation strategies across different stages of the ML development pipeline and compare their effects on predictive performance and fairness in MOUD outcome prediction. This evaluation provides practical guidance for selecting bias mitigation approaches appropriate for different fairness objectives and predictive performance requirements in MOUD treatment settings.  
\end{enumerate}

\section{Related Work}
In this section, we review prior research across three key areas: (1) MOUD treatment outcomes and ML; (2) algorithmic fairness and bias mitigation in ML models. We conclude by identifying gaps in the current literature and describing how our study addresses these challenges. 

\subsection{MOUD Treatment Outcomes and ML}
The effectiveness of MOUD depends not only on treatment initiation but also on sustained engagement over time. Prior literature indicates that fewer than one-third of adults with prescription opioid use disorder ever receive treatment \cite{blanco2013probability}, highlighting substantial barriers at the initiation stage. Among those who receive treatment, retention rates are discouragingly low and vary considerably across different follow-up periods \cite{timko2016retention}. 
Treatment retention, often measured by continued participation over clinically meaningful durations such as 90 or 180 days, has been consistently associated with lower relapse risk, reduced overdose, and improved stability \cite{timko2016retention}. In outpatient MOUD settings, sustained engagement is especially important because treatment occurs in less supervised environments and depends on regular attendance, medication adherence, and continued follow-up over time \cite{baird2023determinants}. 
However, premature treatment discontinuation remains common. Prior studies have shown that a substantial proportion of patients discontinue therapy prematurely within the first weeks or months of treatment, particularly during early phases when patients may still be adjusting to medication, clinic requirements, and psychosocial supports \cite{o2020models,samples2018risk,morgan2021characterizing}. 
Retention outcomes vary widely across treatment settings, patient populations, and medication types, and are shaped by a complex interplay of clinical, behavioral, and structural factors \cite{morgan2018injectable,saloner2017patterns,weinstein2017long}. For example, prior research has documented substantial differences in treatment completion across sociodemographic subgroups, including by race, age, and sex, as well as by social and contextual conditions such as employment status, housing stability, and transportation access \cite{dong2023racial,hasan2021machine,morgan2018injectable,saloner2017patterns}. As a result, identifying patients at elevated risk of premature discontinuation in order to support targeted interventions remains challenging, particularly in outpatient MOUD settings, where engagement is less structured and programs must selectively prioritize limited resources such as care coordination time, counseling and peer-support services, proactive outreach, and follow-up visits. 

ML methods have increasingly been applied to support risk stratification, the process of identifying patients at relatively higher or lower risk for adverse outcomes in order to guide prioritization of care, using routinely collected clinical or administrative data in addiction treatment settings \cite{stafford2022identifying,hasan2021machine,al2024explainable}. This is particularly relevant in the context of MOUD treatment. 
Previous studies have used conventional regression-based methods to identify the significant predictors of OUD-related outcomes \cite{white2009analytic,rice2012model,han2020using}. 
More recently, ML models can flexibly capture complex, non-linear relationships among patient characteristics, treatment factors, and contextual variables, and have increasingly been used to develop predictive tools for OUD risk stratification \cite{krawczyk2021stays,baird2023determinants,stafford2022identifying,welsh2023substance,warren2022using,dong2021identifying}. In real-world settings where data are limited and outcomes are influenced by various factors, ML-based risk stratification offers a pragmatic approach to supporting population-level decision-making and quality improvement efforts. Such models can inform decisions, such as prioritizing care coordination efforts, allocating behavioral health counseling resources, or triggering proactive follow-up for patients at elevated risk of disengagement \cite{hasan2021machine}. 
At the same time, the use of ML-based approaches raises important concerns regarding algorithmic fairness across patient subgroups. If predictive models systematically over- or under-estimate risk for certain populations, risk-informed decision-making may inadvertently reinforce existing disparities in access to care or support services. These concerns motivate the need for careful evaluation of model performance and the incorporation of algorithmic fairness considerations when developing ML models for MOUD treatment outcome prediction. 

\subsection{Algorithmic Fairness and Bias Mitigation}
The growing use of ML has brought increasing attention to \emph{algorithmic bias}, which occurs when AI models consistently produce less accurate or potentially discriminatory outcomes for certain subgroups of people \citep{parikh2019addressing}. 
Although developing separate prediction models for each patient subgroup may appear straightforward, social categories such as race and ethnicity are neither clear-cut nor mutually exclusive \cite{ferryman2018fairness}. Rigidly assigning individuals to a single subgroup and conducting subgroup-specific modeling can therefore misrepresent lived experiences and lead to biased model use. This also risks reinforcing the false assumption that observed disparities reflect biological differences rather than the underlying structural or contextual factors \cite{ferryman2018fairness}. 
Therefore, evaluating differences in model performance across patient subgroups has become an emerging practice for assessing the algorithmic fairness of AI models~\citep{hardt2016equality}. 
Recent studies have established fairness evaluation criteria and model development procedures to quantify and mitigate differences in model performance across population subgroups defined by sensitive attributes such as race and sex \citep{hardt2016equality,agarwal2018reductions}. 
For example, \emph{demographic parity} requires predictions to be statistically independent of sensitive attributes \citep{agarwal2018reductions}. 
\emph{Equal opportunity}~\citep{hardt2016equality} requires equal true positive rates (TPR) across subgroups, while \emph{equalized odds} \citep{hardt2016equality} adds an additional constraint that demands equal false positive rates (FPR). Both require that outcomes be conditionally independent of sensitive attributes given the true outcome. All of these metrics aim to enforce performance parity across subgroups. 
Since there is no consensus on a single metric or criterion for assessing algorithmic fairness, the choice of fairness definition often depends on the specific characteristics of the application.

Various approaches have been developed to achieve fairness in ML, which are generally classified into three categories: pre-processing, in-processing, and post-processing methods. 
Pre-processing methods focus on modifying input data to reduce biases before training, including techniques like resampling, adding new data, or adjusting labels \citep{feldman2015certifying,jiang2020identifying,abernethy2022active}. 
Post-processing methods adjust model predictions after training to meet fairness objectives, often by modifying decision thresholds or outcomes for specific subgroups \citep{kim2019multiaccuracy}. 
In-processing methods incorporate fairness constraints or objectives directly into the learning algorithm during training, penalizing the learning of discriminatory features and enabling a balance between fairness and predictive performance. Examples include adversarial training, regularization, adaptive weighting, or fairness constraints on representations \citep{agarwal2018reductions,fish2016confidence,zhang2018mitigating}. 

In healthcare, algorithmic bias in ML models can arise from multiple sources, including structural inequities in access to care, differences in data completeness or quality across populations, and the use of proxy variables or shortcuts that reflect social and contextual factors rather than true clinical risk \cite{obermeyer2019dissecting,leslie2021does,chen2023algorithmic}. As a result, models optimized for overall predictive performance may produce unfair outputs across patient subgroups, potentially reinforcing existing disparities. Prior research has documented fairness concerns in a range of healthcare applications, including risk prediction, disease screening, and resource allocation, demonstrating that models with acceptable overall performance may still systematically disadvantage certain population subgroups \cite{obermeyer2019dissecting,chen2023algorithmic,seyyed2021underdiagnosis,li2023evaluating}. 
In the context of MOUD, existing literature suggests that structural and contextual factors play an important role in patients’ sustained engagement in treatment \citep{dong2023racial,hasan2021machine}. Community-level context, including the availability of treatment providers, transportation infrastructure, and the broader socioeconomic environment, can further influence treatment adherence and outcomes \citep{kennedy2022factors}. When such variations and patterns are encoded in routinely collected clinical or administrative data, they may also be implicitly learned by ML models, causing unfair and biased outcomes across patient subgroups \cite{chen2023algorithmic}. 
However, fairness considerations have received limited attention in ML models predicting MOUD treatment outcomes. As predictive models are increasingly used to support risk stratification and care prioritization, it is important to systematically examine whether these models perform fairly across patient subgroups and to assess whether bias mitigation strategies can reduce observed disparities. 
The choice of fairness criteria should therefore be guided by the intended use of the ML model and the potential consequences of different types of prediction errors. 
In this study, we focus on error-rate–based fairness metrics, as the prediction model is intended to support risk stratification and care prioritization in MOUD treatment settings. In this context, unfair false negative rates (FNR) may lead to missed opportunities for early intervention, while unfair FPR may result in unnecessary monitoring or resource allocation for certain patient subgroups. Evaluating TPR and FPR parity therefore directly reflects whether patients from different subgroups have equitable access to follow-up and support when risk-based decisions are made. Such fairness metrics have been widely used in prior healthcare literature where models inform disease screening, monitoring, or resource allocation decisions \cite{seyyed2021underdiagnosis,vaidya2024demographic}. 


\section{Methods}
\begin{figure}[tbp]
\centering
\includegraphics[width=0.98\textwidth]{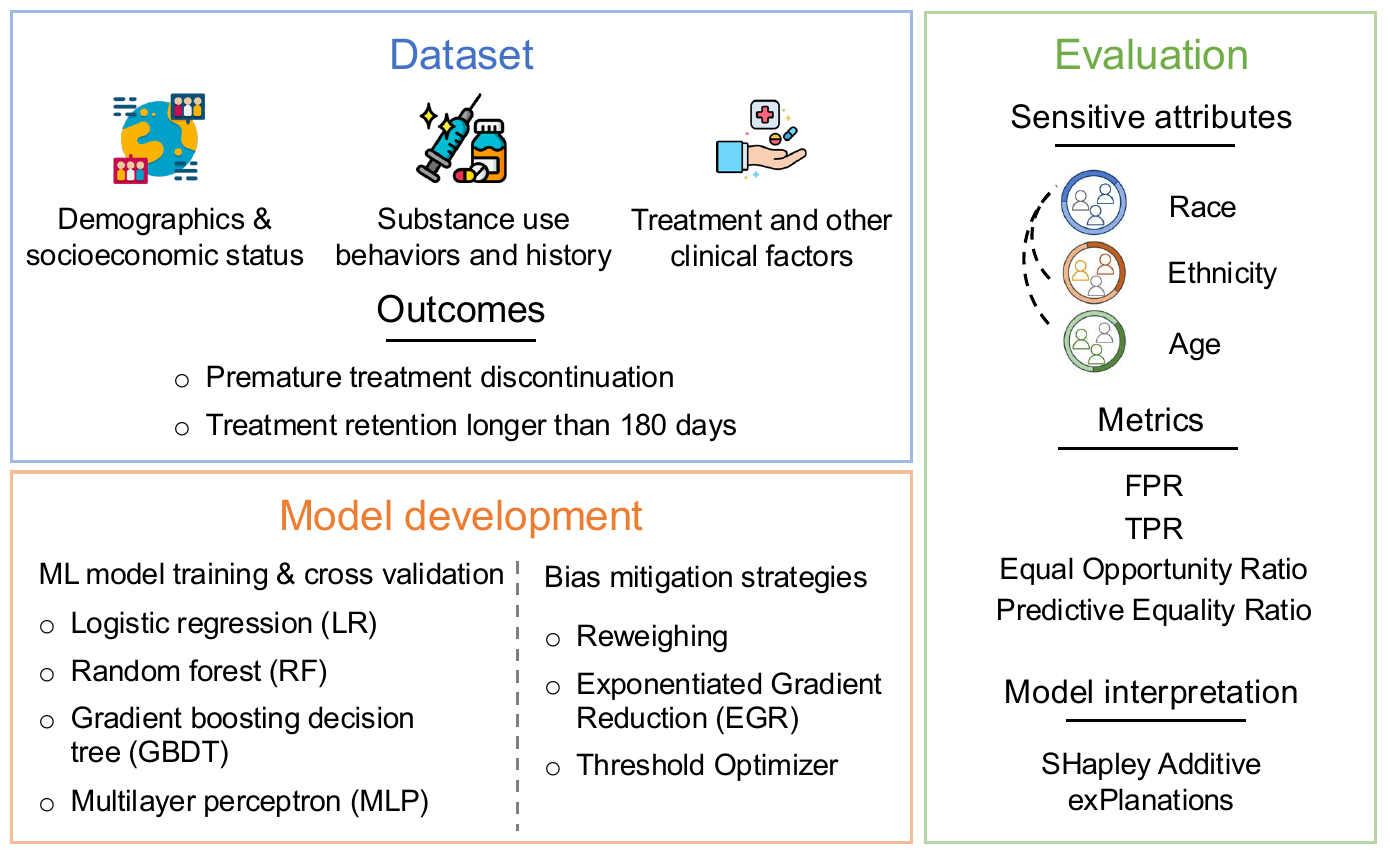}
\caption{Overall pipeline.}
\label{fig:pipeline}
\end{figure}

This retrospective cross-sectional study used data from the U.S. Substance Abuse and Mental Health Services Administration's (SAMHSA) Treatment Episode Data Set - Discharges (TEDS-D) \cite{SAMHSA2021}, focusing on adult cases receiving MOUD in outpatient treatment facilities between 2015 and 2019. 
In this study, we focused on two clinically important and complementary treatment outcomes in MOUD programs: premature treatment discontinuation and treatment retention exceeding 180 days. Treatment retention is a key indicator of successful treatment engagement, whereas premature treatment discontinuation is associated with poor treatment outcomes and an increased risk of relapse, overdose, and mortality. 
Our objective was not only to develop predictive models for these outcomes, but also to systematically examine how model performance varies across patient subgroups defined by sociodemographic characteristics such as race, ethnicity, and age, which are well-documented attributes often linked to differences in treatment access and outcomes. By treating these variables as sensitive attributes, we conducted a comprehensive assessment of whether prediction accuracy and behavior differed across these subgroups, and implemented bias mitigation strategies to improve the fairness of model performance across the patient sociodemographic subgroups. 
The overall process of our pipeline is shown in Figure~\ref{fig:pipeline}. 

\subsection{Data}

\subsubsection{Data Sources and Study Population}
This study utilized the 2015–2019 TEDS-D dataset \cite{SAMHSA2021}, a national data system of annual discharges from substance use treatment facilities licensed or certified by Single State Agencies and receiving federal funding. 
TEDS-D provides episode-level information on discharges from substance use treatment services. 
The dataset includes individuals aged 12 and older and covers a wide range of demographic, clinical and treatment-related variables, as well as substance use characteristics and history for individuals admitted to treatment programs. Importantly, each record in the dataset corresponds to a treatment episode rather than a unique individual. TEDS-D enables large-scale, population-level analyses of substance use treatment outcomes and is commonly used to examine patterns in service utilization, treatment completion, and other discharge-related outcomes across a wide range of treatment settings in the United States. 


We derived cohort based on specific inclusion and exclusion criteria to ensure the robustness and consistency of the results. 
First, the study cohort was restricted to adult patients (aged 18 and older) receiving MOUD, which was defined as the documented use of methadone, buprenorphine, or naltrexone in the patient's treatment plan at the time of admission \cite{stahler2021racial}. 
Restricting the cohort to adults ensured demographic consistency and allowed us to focus on a population with well-documented treatment trajectories and distinct clinical needs. 
To focus on opioid-related treatment outcomes, we further restricted the cohort to discharges where heroin or other opioids/synthetics were documented as the primary, secondary, or tertiary substance at admission, allowing us to specifically target patients with opioid use \cite{stafford2022identifying}. 
We further restricted the cohort to individuals receiving treatment in outpatient settings to ensure consistency in treatment environments \cite{krawczyk2021stays}. Outpatient settings present a substantially more challenging prediction task than mixed service settings, as treatment engagement is less structured and outcomes are influenced by dynamic and often unobserved factors. In such settings, ML may provide useful support for identifying patterns that are otherwise difficult to capture. 
Finally, we excluded discharges classified as transfers to other treatment programs or facilities, preventing interruptions in treatment continuity from confounding the analysis and ensuring that the treatment outcomes evaluated correspond to a single, uninterrupted episode of care. 
A total of 366,825 patients satisfied the respective inclusion/exclusion criteria for the model development. 
Figure \ref{fig:cohort_variable} presents the complete cohort and variable selection process. 
In addition, to assess the robustness of our analysis, we conducted additional experiments using the 2020–2023 TEDS-D data following the same study design and analysis pipeline. The results are presented in Appendix~\ref{results_20_23}.

\subsubsection{Outcome Measures}
This study examined two outcomes of interest: premature discontinuation of MOUD treatment and treatment retention beyond 180 days, defined as follows: 
\begin{itemize}
    \item[(1)] \textbf{Premature Discontinuation}: A binary outcome variable coded as positive for individuals who discontinue treatment prematurely due to potentially preventable attrition, that is, cases in which people ``chose'' not to complete treatment. Specifically, this category included discharge reasons such as dropout for unknown reasons, loss to follow-up, failure to return from leave, or discharge for administrative purposes following a prolonged absence from treatment. These discharge reasons represent situations in which treatment discontinuation may be amenable to clinical or programmatic intervention (e.g., counseling, outreach, or adjustments to the treatment plan). The variable was coded as negative for all other discharge reasons, including treatment completion; termination by the facility due to non-compliance or violations of program rules, laws, or policies; incarceration (including jail, prison, house confinement, or release to or from the courts); and death. These discharge reasons were not considered premature discontinuation because they primarily reflect institutional decisions, external circumstances, or other reasons for treatment termination rather than patient disengagement from treatment, making them less directly amenable to interventions aimed at improving patient retention. Transfers to another treatment program or facility were excluded \cite{stafford2022identifying}. This outcome was derived from the ``reason for discharge'' variable in the original dataset and was based solely on the recorded discharge reason, consistent with prior work \cite{stafford2022identifying}. 
    
    \item[(2)] \textbf{Length of Stay (LOS) $>$ 180 Days}: A binary variable measuring treatment retention or sustained engagement beyond 180 days. The variable was coded as positive if the treatment episode lasted more than 180 days and negative otherwise, aligned with established treatment cascade metrics for MOUD continuity \cite{krawczyk2021stays,williams2018developing,askari2020medication}. The 180-day threshold was selected based on prior studies demonstrating that patients who discontinue MOUD within the first six months are at substantially higher risk of relapse and adverse outcomes compared to those retained beyond this period \cite{jones2022receipt,williams2019development}. This cutoff has been widely used in MOUD research and clinical trials as a clinically meaningful marker of sustained treatment engagement \cite{jones2022receipt,lopez2024predicting}. Accordingly, retention beyond 180 days identifies individuals who have achieved a level of continuity of treatment associated with improved outcomes, whereas shorter treatment episodes may indicate elevated risk and the need for additional support. This outcome was derived from the ``length of stay in treatment (days)'' variable in the original dataset and was defined exclusively based on treatment duration over a clinically meaningful period. 
\end{itemize}
Because the TEDS-D dataset does not contain a single variable that simultaneously captures both treatment duration and the reason for discharge, we operationalized treatment engagement using these two complementary outcomes. Both outcomes were non-missing for all treatment episodes, ensuring complete data availability for evaluation and analysis.

\subsubsection{Predictor Variables}
To predict treatment outcomes, we included a range of variables collected at admission, spanning demographic characteristics (e.g., race, ethnicity, sex, and age)~\cite{lappan2020dropout,krawczyk2021stays}, socioeconomic factors (e.g., employment status, education, veteran status, living situation, geographic region, and arrest history)~\cite{gautam2020machine,binswanger2013mortality,sayre2002determining,sugarman2020interventions}, substance use behaviors (e.g., number of substance used)~\cite{stafford2022identifying}, self-reported substances (e.g., heroin, alcohol, inhalants, and marijuana/hashish)~\cite{stafford2022identifying}, other clinical factors (e.g., presence of psychological problems)~\cite{blanco2019management,sayre2002determining}, and treatment-related factors (e.g., source of referral, service setting, and prior treatment history)~\cite{brorson2013drop,stafford2022identifying}. 
Specifically, we first excluded variables that were not relevant or applicable to the majority of the study population and therefore had no recorded values for more than 90\% of individuals. 
And we only considered the variables collected at admission. 
We also derived several variables to better capture key patterns, including the highest frequency of non-medical opioid use recorded among participants, a binary indicator for any current heroin use, and four distinct binary variables indicating whether participants had ever used substances within the broader categories of stimulants, hallucinogens, sedatives, or tranquilizers. 
Finally, among the remaining candidate variables, those with more than 20\% missing values were excluded. 
In total, 27 variables were included, 2 were target variables (i.e., retention and premature discontinuation), leaving 25 variables for use as predictors.

\subsection{Prediction Models}
In this study, we assessed algorithmic fairness in the prediction of MOUD retention and premature discontinuation across four ML models: LR, RF, GBDT, and MLP. These models were selected based on their widespread use in clinical prediction tasks and their complementary strengths in handling structured, tabular healthcare data \cite{zhang2019predictive,fernandez2021random,nusinovici2020logistic}. 
LR is a widely used baseline model for binary classification problems. t assumes a linear relationship between predictors and the log-odds of the outcome, making it particularly useful for establishing foundational performance benchmarks. 
RF is an ensemble-based method that constructs multiple decision trees and aggregates their outputs to improve accuracy and robustness. Its ability to capture non-linear interactions and rank variable importance has made it a standard model in medical applications involving complex clinical features. 
GBDT is a boosting-based ensemble method that iteratively reduces errors made by previous learners, which enhances its capacity to model subtle and complex relationships. GBDT models are often considered state-of-the-art in structured data scenarios due to their high predictive accuracy and versatility in handling heterogeneous variables. 
MLP is a fully connected feedforward neural network that can learn complex, non-linear patterns in data. Although MLPs typically require more extensive tuning and are less interpretable than tree-based models, they have shown competitive performance in structured clinical prediction when sufficient data are available. 
By leveraging both traditional and neural network–based approaches, we aim to provide a comprehensive evaluation of predictive performance and subgroup-level reliability across a range of ML models.

\subsection{Empirical Analysis}
Hyperparameters were optimized separately for each model using the procedures described below. Data were randomly partitioned into 70\% for training and 30\% for testing using stratified sampling. The training partition was further divided into 80\% for training and 20\% for validation. The validation set was used for hyperparameter tuning for the MLP model, whereas LR, RF, and GBDT used five-fold stratified cross-validation on the training subset. 
All variables were categorical variables and were converted to one-hot encoding before being fed into ML models. 
Independent models were trained and optimized for premature treatment discontinuation and treatment retention exceeding 180 days, as these outcomes are defined by different mechanisms. 
Models were selected based on the area under the receiver operating characteristic (AUC), with results averaged over 10 independent runs using different random seeds to account for variability. 
Best hyperparameter configurations for each model and task are summarized in Appendix~\ref{hyperparameter_tuning} Table~\ref{tab:hyperparameter}. 

\subsubsection{Subgroup-Level Model Performance Evaluation}
To evaluate how predictive performance varies across patient subgroups, we analyzed model predictions based on sociodemographic characteristics available in the dataset as sensitive attributes, with a focus on fields related to recorded race, ethnicity, and age. 
The original TEDS-D dataset includes nine racial categories, but due to small sample sizes in many of these subgroups, we limited our subgroup analysis to White and Black/African American patients. Data samples from other racial categories were excluded from our evaluation. Two ethnicity subgroups, Non-Hispanic and Hispanic, were considered. Patients were also stratified into three age subgroups: 18-24 (young adults), 25-54 (middle-aged adults), and over 55 years (older adults). 

To examine potential differences in predictive behavior, we used key classification metrics derived from the confusion matrix: true positive rate (TPR or sensitivity), false positive rate (FPR), false negative rate (FNR) and true negative rate (TNR), as listed in Table~\ref{tab:metrics}. 
These metrics are particularly relevant to clinical decision-making and resource allocation. Since FNR and TNR can be directly derived from TPR and FPR, we primarily focused our analysis on TPR and FPR. 
High FPR values indicate that patients who would have remained in treatment are incorrectly flagged as being at risk of early discontinuation, which can lead to unnecessary interventions, misallocated resources, and negative perceptions. Low TPR values, on the other hand, suggest that many patients who are truly at risk are not identified in time, limiting the opportunity for timely intervention and support. In the task of predicting whether patients will remain in treatment for more than 180 days, the same principles apply: accurate identification helps ensure that extended-care resources are directed to those who truly need them, while minimizing both missed opportunities and inefficient resource use. 

\begin{table}[htb]
\caption{Confusion matrix and derived metrics}\label{tab:metrics}%
\begin{tabular}{@{}lll@{}}
\toprule
  & Predicted - Positive  & Predicted - Negative  \\
\midrule
\multirow{2}{*}{Actual - Positive} 
  &  True Positive (TP)   &  False Negative (FN)  \\
  &  $TPR=\frac{TP}{TP+FN}$  &  $FNR=1-TPR$  \\
\midrule
\multirow{2}{*}{Actual - Negative} 
  &  False Positive (FP)   &  True Negative (TN)  \\
  &  $FPR=\frac{FP}{FP+TN}$  &  $TNR=1-FPR$  \\
\botrule
\end{tabular}
\end{table}

The above metrics also align with two widely used subgroup fairness criteria: \emph{Equal Opportunity} and \emph{Predictive Equality} \cite{hardt2016equality}. 
Equal opportunity ensures that TPRs remain consistent across subgroups defined by a sensitive attribute. It evaluates how well a model correctly identifies positive cases, ensuring that individuals with a positive ground truth outcome (e.g., premature discontinuation or retention longer than 180 days) have an equal likelihood of being predicted as such, regardless of their sensitive attribute. The formal definition in the binary-subgroup case is: 
\begin{equation}
P(\hat{Y}=1|Y=1,A=0) = P(\hat{Y}=1|Y=1,A=1)
\end{equation}
where $Y \in \{0,1\}$ denotes the actual outcome, $\hat{Y} \in \{0,1\}$ is the model's predicted outcome, $A \in \{0,1\}$ represents the sensitive attribute (e.g., race, age). 
In practice, however, sensitive attributes often include more than two subgroups—for instance, multiple racial/ethnic subgroups or age categories. In such cases, equal opportunity can be generalized to require that the TPRs be similar across all subgroups $\mathcal{A} = \{a_1,a_2,\ldots,a_N\}$, where $N$ denotes the total number of subgroups. 
To quantify the extent to which equal opportunity is satisfied across multiple subgroups, we measure the \emph{Equal Opportunity Ratio} (EOR). This metric compares the TPRs between subgroups defined by the sensitive attribute $A$. Formally, it is defined as: 
\begin{equation}
\text{EOR} = \frac{\min_{a_i \in \mathcal{A}} P(\hat{Y}=1 \mid Y=1, A=a_i)}{\max_{a_i \in \mathcal{A}} P(\hat{Y}=1 \mid Y=1, A=a_i)}, 
\end{equation}
where $\mathcal{A}$ is the set of all subgroups defined by sensitive attribute $A$. This ratio captures the worst-case gap in TPRs across subgroups. A value closer to 1 indicates that the model identifies positive cases at a similar rate for all subgroups, aligning with the principle of equal opportunity. Lower values signal greater gaps, meaning the model favors certain subgroups over others in predicting positive outcomes. 

Predictive Equality focuses on ensuring that FPRs are consistent across subgroups defined by a sensitive attribute. It ensures that individuals who do not actually exhibit the outcome of interest (e.g., those who do not discontinue MOUD treatment prematurely, or those who do not stay in treatment longer than 180 days) are not disproportionately predicted as high-risk simply because they belong to a particular subgroup. Formally, the definition in the binary-subgroup setting is: 
\begin{equation}
P(\hat{Y}=1|Y=0,A=0) = P(\hat{Y}=1|Y=0,A=1)
\end{equation}
Similarly, we use the \emph{Predictive Equality Ratio} (PER) to quantify the extent to which predictive equality is satisfied: 
\begin{equation}
\text{PER} = \frac{\min_{a_i \in \mathcal{A}} P(\hat{Y}=1 \mid Y=0, A=a_i)}{\max_{a_i \in \mathcal{A}} P(\hat{Y}=1 \mid Y=0, A=a_i)}. 
\end{equation}
This ratio captures the worst-case gap in FPRs across subgroups. A PER close to 1 indicates that the model is making false positive errors at a similar rate across all subgroups, which aligns with the fairness goal of avoiding disproportionate over-prediction for certain population subgroups. Values far from 1 highlight discrepancies that could lead to unequal burdens, such as unnecessary interventions or stigma disproportionately affecting specific subgroups. 

In addition, we also evaluate Equalized Odds, which jointly measures gaps in TPRs and FPRs across patient subgroups. 
To quantify deviations from this criterion, we use the Equalized Odds Difference (EOD), which is defined as the maximum of the TPR difference and the FPR difference across subgroups, as follows: 
\begin{align}
\Delta_{\mathrm{TPR}}
&=
\max_{a_i \in \mathcal{A}}
P(\hat{Y}=1 \mid Y=1, A=a_i)
-
\min_{a_i \in \mathcal{A}}
P(\hat{Y}=1 \mid Y=1, A=a_i),
\\
\Delta_{\mathrm{FPR}}
&=
\max_{a_i \in \mathcal{A}}
P(\hat{Y}=1 \mid Y=0, A=a_i)
-
\min_{a_i \in \mathcal{A}}
P(\hat{Y}=1 \mid Y=0, A=a_i),
\\
\Delta_{\mathrm{EOD}}
&=
\max\left(\Delta_{\mathrm{TPR}}, \Delta_{\mathrm{FPR}}\right).
\end{align}

To assess the fairness of the ML models, we compared TPRs and FPRs across patient subgroups, as these metrics directly quantify gaps in prediction outcomes at the selected decision threshold. We also evaluated subgroup AUROC as a complementary, threshold-independent measure of overall discriminative performance, with the results presented in Appendix~\ref{subgruop_AUROC}. 

\subsubsection{Model Explanations}

To better understand model behavior and enhance transparency, we computed SHAP values as a means of interpreting feature contributions. SHAP (SHapley Additive exPlanations) provides a unified, theoretically grounded approach to feature attribution by quantifying the marginal contribution of each input feature to the model’s output \cite{lundberg2017unified}. It is model-agnostic and rooted in cooperative game theory, building on the concept of Shapley values, where each input feature is considered a ``player'' in a game contributing to the model’s prediction. 
Formally, SHAP values are based on additive feature attribution methods, which assume the explanation model is a linear function of binary variables: 
\[
g(z') = \phi_0 + \sum_{i=1}^{M} \phi_i z'_i
\]
where \( z' \in \{0, 1\}^M \) represents the simplified input, a binary vector indicating the presence or absence of feature $i$ in the explanation, \( M \) is the number of input features, \( \phi_0 \) is the expected model output when no features are present (base value), and \( \phi_i \in \mathbb{R} \) represents the contribution (SHAP value) of feature \( i \). 
SHAP values capture how much each feature contributes to pushing the model’s prediction above or below the baseline (i.e., the expected output across all instances). A positive SHAP value indicates that a feature increases the model’s predicted probability of the target outcome (e.g., premature discontinuation or retention longer than 180 days), while a negative SHAP value suggests the opposite. 
Importantly, SHAP values are computed for each instance, allowing for both local explanations (individual-level insights) and global explanations (aggregated feature importance across the dataset).

\subsection{Bias Mitigation Strategies}

We systematically assessed the effectiveness of several bias mitigation strategies to understand their ability to reduce performance inconsistencies and improve algorithmic fairness across patient subgroups in ML models. 
To provide a comprehensive evaluation, we selected representative techniques that operate at three different stages of the ML development pipeline: pre-processing, in-processing, and post-processing. 
These methods are widely recognized in the ML fairness literature, and have been shown to be effective in improving fairness across a range of real-world applications \cite{chen2024unmasking}. 
Specifically, we evaluated three approaches. 

\subsubsection{Reweighing} 
Reweighing \cite{kamiran2012data} is a pre-processing technique that assigns different weights to training instances based on their subgroup membership and outcome label. The goal is to adjust the influence of each sample during training to create a more balanced dataset. By increasing the weight of subgroups with limited data, this method encourages the model to learn from a more balanced representation of the population. This helps mitigate learning patterns that disproportionately favor certain subgroups, without modifying the model architecture itself. 
Specifically, let the training examples consist of triples $(X,A,Y)$, where $X \in \mathcal{X}$ is a feature vector, $A \in \mathcal{A}$ is the protected attribute (e.g., race) and $Y \in {0,1}$ is the outcome label. Each training instance is assigned a weight based on its subgroup membership $A$ and outcome label $Y$: 
\[
w(a,y) = \frac{P(A=a)P(Y=y)}{P(A=a,Y=y)}.
\]
These weights reflect the ratio between the expected frequency of each $(A,Y)$ combination under statistical independence and its observed frequency in the data. Underrepresented subgroup–outcome combinations receive larger weights, while overrepresented combinations receive smaller weights. The resulting weights are incorporated into the model’s loss function during training. 

\subsubsection{Exponentiated Gradient Reduction (EGR)} 
Exponentiated Gradient Reduction (EGR) \cite{agarwal2018reductions} is a model-specific in-processing technique that modifies the learning algorithm by incorporating fairness constraints during model training. It works by adjusting how the model selects decision boundaries, seeking a balance between accuracy and fairness. Specifically, EGR treats fairness as an optimization constraint—ensuring that prediction outcomes do not differ substantially across subgroups—while still aiming to minimize prediction error. 
Let $f$ denote the prediction model, $L(f)$ denote the empirical prediction loss, and $g_k(f)$ denote $k$ fairness constraint functions that quantify gaps in error rates (e.g., TPR or FPR) across subgroups defined by $A$, EGR formulates model training as a constrained optimization problem: 
\[
\min_{f} L(f) \quad \text{s.t.} \quad |g_k(f)| \leq \epsilon, \forall k,
\]
where $\epsilon$ specifies the allowable tolerance for fairness violations. This constrained problem is solved via a Lagrangian formulation, in which fairness constraints are incorporated through dual variables. At each iteration, training samples are reweighted according to the current dual variables, and a base classifier is trained on the reweighted data. The procedure alternates between minimizing prediction loss and adjusting weights to reduce fairness gaps. 
EGR is a model-agnostic approach that integrates directly into the training loop and can be applied to a wide range of predictive models. The final predictor produced by EGR is a randomized mixture of classifiers obtained across iterations, which jointly balances predictive accuracy and fairness constraints. 

\subsubsection{Threshold Optimizer} 
Threshold Optimizer \cite{hardt2016equality} is a post-processing method that adjusts the decision thresholds applied to predicted risk scores for different subgroups in order to satisfy a specified fairness criterion after the model has been trained. By customizing thresholds per subgroup, the model can produce more consistent outputs across subgroups, helping to address fairness concerns without retraining or altering the underlying data. 
Specifically, it learns subgroup-specific decision thresholds for converting continuous risk scores $\hat{s} = f(X)\in[0,1]$ into binary predictions $\hat{Y}$. Using the protected attribute $A$, a threshold $\tau_a$ is chosen for each subgroup $a\in A$ so that: 
\[
\hat{Y} = \mathbb{I}\big[\hat{s} \ge \tau_a\big],
\]
where $\mathbb{I}[\cdot]$ is the indicator function. In practice, the Threshold Optimizer can also produce a randomized classifier (a mixture of thresholds) to achieve fairness constraints exactly, particularly when deterministic thresholds cannot satisfy the requirements. 
It selects thresholds to satisfy the specified fairness constraints (e.g., equal opportunity and predictive equality) while minimizing a classification error metric in predictive performance. Formally, this can be expressed as:
\[
\min_{\{\tau_a\}} \; \text{Error}(\hat{Y}(\{\tau_a\}))
\quad
\text{s.t.}
\quad
|\mu_k(\hat{Y})| \le \epsilon,\;\forall k,
\]
where $\mu_k(\hat{Y})$ are fairness gap functions (e.g., TPR or FPR differences across subgroups) and $\epsilon$ is an allowable tolerance. From a practitioner perspective, this approach can be useful because it allows organizations to improve fairness in downstream decisions without retraining models or altering established workflows, which may be particularly valuable in resource-constrained clinical environments. 
In our study, the three approaches were optimized on validation data using Fairlearn~\cite{bird2020fairlearn} or AIF360~\cite{aif360-oct-2018}. 

\setlength{\tabcolsep}{3.3pt}
\begin{table}[htbp]
\caption{Characteristics of cohorts ($N = 366,825$) presented as the number of episodes, with percentages indicating their proportion within the entire cohort (not within each outcome category).}
\label{tab:dataset}
\centering
\footnotesize
\begin{tabular*}{\textwidth}{@{}llcc|cc@{}}
\toprule
& & \multicolumn{2}{c}{\textbf{Premature Discontinuation}} & \multicolumn{2}{c}{\textbf{Treatment Retention}} \\

\cmidrule(lr){3-4} \cmidrule(lr){5-6} 

& & Discontinuation & Completion 
& $>$ 180 days & $\leq$ 180 days \\

\midrule
\multirow{2}{*}{Sex} 
& Male 
& 115,613 (31.52\%) & 97,178 (26.49\%) 
& 81,857 (22.32\%) & 130,934 (35.69\%) \\
& Female 
& 83,008 (22.63\%) & 71,026 (19.36\%) 
& 59,910 (16.33\%) & 94,124 (25.66\%) \\

\midrule
\multirow{2}{*}{Race} 
& White 
& 163,842 (44.66\%) & 145,169 (39.48\%) 
& 116,872 (31.86\%) & 192,139 (52.28\%) \\
& Black 
& 34,779 (9.48\%) & 23,035 (6.28\%) 
& 24,895 (6.79\%) & 32,919 (8.97\%) \\

\midrule
\multirow{2}{*}{Ethnicity} 
& Non-Hispanic 
& 182,285 (49.69\%) & 157,103 (42.83\%) 
& 130,078 (35.46\%) & 209,310 (57.06\%) \\
& Hispanic 
& 16,336 (4.45\%) & 11,101 (3.03\%) 
& 11,689 (3.19\%) & 15,748 (4.29\%) \\

\midrule
\multirow{3}{*}{Age} 
& 18-24 
& 19,569 (5.33\%) & 16,526 (4.51\%) 
& 12,480 (3.40\%) & 23,615 (6.44\%) \\
& 25-54 
& 156,682 (42.71\%) & 135,436 (36.92\%) 
& 109,938 (29.97\%) & 182,179 (49.66\%) \\
& $\geq$ 55 
& 22,370 (6.10\%) & 16,243 (4.43\%) 
& 19,349 (5.27\%) & 19,264 (5.25\%) \\

\midrule
\multirow{5}{*}{Education} 
& Grade 8 or less 
& 9,776 (2.67\%) & 6,714 (1.83\%) 
& 6,545 (1.78\%) & 9,945 (2.71\%) \\
& Grades 9-11 
& 40,862 (11.14\%) & 33,146 (9.04\%) 
& 29,036 (7.92\%) & 44,972 (12.26\%) \\
& Grade 12 
& 95,113 (25.93\%) & 81,765 (22.29\%) 
& 66,775 (18.20\%) & 110,103 (30.02\%) \\
& College or more 
& 50,160 (13.67\%) & 41,467 (11.31\%) 
& 36,393 (9.92\%) & 55,234 (15.06\%) \\
& Unknown 
& 2,710 (0.74\%) & 5,112 (1.39\%) 
& 3,018 (0.82\%) & 4,804 (1.31\%)  \\

\midrule
\multirow{4}{*}{Employment} 
& Full-time 
& 30,178 (8.23\%) & 28,396 (7.74\%) 
& 23,952 (6.53\%) & 34,622 (9.44\%) \\
& Part-time 
& 16,123 (4.40\%) & 14,612 (3.98\%) 
& 12,431 (3.39\%) & 18,304 (4.99\%) \\
& Unemployed 
& 148,021 (40.35\%) & 119,799 (32.66\%) 
& 102,118 (27.84\%) & 165,702 (45.17\%) \\
& Unknown 
& 4,299 (1.17\%) & 5,397 (1.47\%) 
& 3,266 (0.89\%) & 6,430 (1.75\%) \\

\midrule
\multirow{4}{*}{Housing} 
& Independent 
& 153,761 (41.92\%) & 132,941 (36.24\%) 
& 114,551 (31.23\%) & 172,151 (46.93\%) \\
& Dependent 
& 25,535 (6.96\%) & 19,585 (5.34\%) 
& 16,105 (4.39\%) & 29,015 (7.91\%) \\
& Homeless 
& 17,032 (4.64\%) & 11,112 (3.03\%) 
& 8,720 (2.38\%) & 19,424 (5.30\%) \\
& Unknown 
& 2,293 (0.63\%) & 4,566 (1.24\%) 
& 2,391 (0.65\%) & 4,468 (1.22\%) \\

\midrule
\multirow{4}{*}{Region} 
& Northeast 
& 99,009 (26.99\%) & 104,350 (28.45\%)
& 79,608 (21.70\%) & 123,751 (33.74\%) \\
& Midwest 
& 30,495 (8.31\%) & 21,196 (5.78\%) 
& 18,537 (5.05\%) & 33,154 (9.04\%) \\
& South 
& 17,614 (4.80\%) & 20,197 (5.51\%) 
& 9,393 (2.56\%) & 28,418 (7.75\%) \\
& West 
& 51,380 (14.01\%) & 22,430 (6.11\%) 
& 34,138 (9.31\%) & 39,672 (10.81\%) \\
& U.S. territories  
& 123 (0.03\%) & 31 (0.01\%) 
& 91 (0.02\%) & 63 (0.02\%) \\

\midrule
\multirow{3}{*}{Arrest\footnotemark[1]} 
& Arrest 
& 9,609 (2.62\%) & 10,190 (2.78\%) 
& 6,147 (1.68\%) & 13,652 (3.72\%) \\
& No arrest 
& 187,259 (51.05\%) & 155,115 (42.29\%) 
& 132,700 (36.18\%) & 209,674 (57.16\%) \\
& Unknown 
& 1,753 (0.48\%) & 2,899 (0.79\%) 
& 2,920 (0.80\%) & 1,732 (0.47\%) \\

\midrule
\multirow{3}{*}{Prior Tx\footnotemark[2]} 
& One or more 
& 145,219 (39.59\%) & 126,297 (34.43\%) 
& 105,161 (28.67\%) & 166,355 (45.35\%) \\
& No prior Tx 
& 50,941 (13.89\%) & 37,765 (10.30\%) 
& 33,418 (9.11\%) & 55,288 (15.07\%) \\
& Unknown 
& 2,461 (0.67\%) & 4,142 (1.13\%) 
& 3,188 (0.87\%) & 3,415 (0.93\%) \\

\midrule
\multirow{3}{*}{Psych\footnotemark[3]} 
& Psych 
& 77,786 (21.21\%) & 74,807 (20.39\%) 
& 53,599 (14.61\%) & 98,994 (26.99\%) \\
& No psych 
& 107,025 (29.18\%) & 79,042 (21.55\%) 
& 76,215 (20.78\%) & 109,852 (29.95\%) \\
& Unknown 
& 13,810 (3.76\%) & 14,355 (3.91\%) 
& 11,953 (3.26\%) & 16,212 (4.42\%) \\

\midrule
\multirow{1}{*}{Injection\footnotemark[4]} 
& Injection use 
& 108,649 (29.62\%) & 91,614 (24.97\%) 
& 75,440 (20.57\%) & 124,823 (34.03\%) \\

\midrule
\multirow{8}{*}{\makecell{Other \\ Substances} } 
& Heroin 
& 163,042 (44.45\%) & 133,990 (36.53\%) 
& 115,034 (31.36\%) & 181,998 (49.61\%) \\
& Alcohol 
& 22,959 (6.26\%) & 22,502 (6.13\%) 
& 15,464 (4.22\%) & 29,997 (8.18\%) \\
& Inhalant 
& 56 (0.02\%) & 98 (0.03\%) 
& 49 (0.01\%) & 105 (0.03\%) \\
& Marijuana 
& 35,877 (9.78\%) & 33,938 (9.25\%) 
& 24,320 (6.63\%) & 45,495 (12.40\%) \\
& Stimulant 
& 67,514 (18.40\%) & 55,523 (15.14\%) 
& 40,951 (11.16\%) & 82,086 (22.38\%) \\
& Tranquilizer 
& 12,205 (3.33\%) & 13,707 (3.74\%) 
& 9,291 (2.53\%) & 16,621 (4.53\%) \\
& Sedative 
& 835 (0.23\%) & 842 (0.23\%) 
& 612 (0.17\%) & 1,065 (0.29\%) \\
& Hallucinogen 
& 4,298 (1.17\%) & 4,377 (1.19\%) 
& 3,149 (0.86\%) & 5,526 (1.51\%) \\

\midrule
& \textbf{Total} 
& 198,621 (54.15\%) & 168,204 (45.85\%) 
& 141,767 (38.65\%) & 225,058 (61.35\%) \\
\bottomrule
\end{tabular*}
\footnotetext[1]{Arrest: arrests in past 30 days prior to admission.}
\footnotetext[2]{Tx stands for treatment.}
\footnotetext[3]{Psych: has co-occurring mental or behavioral health disorders.}
\footnotetext[4]{Injection: have drugs used by injection.}
\end{table}
\begin{figure}[tbp]
\centering
\includegraphics[width=0.72\textwidth]{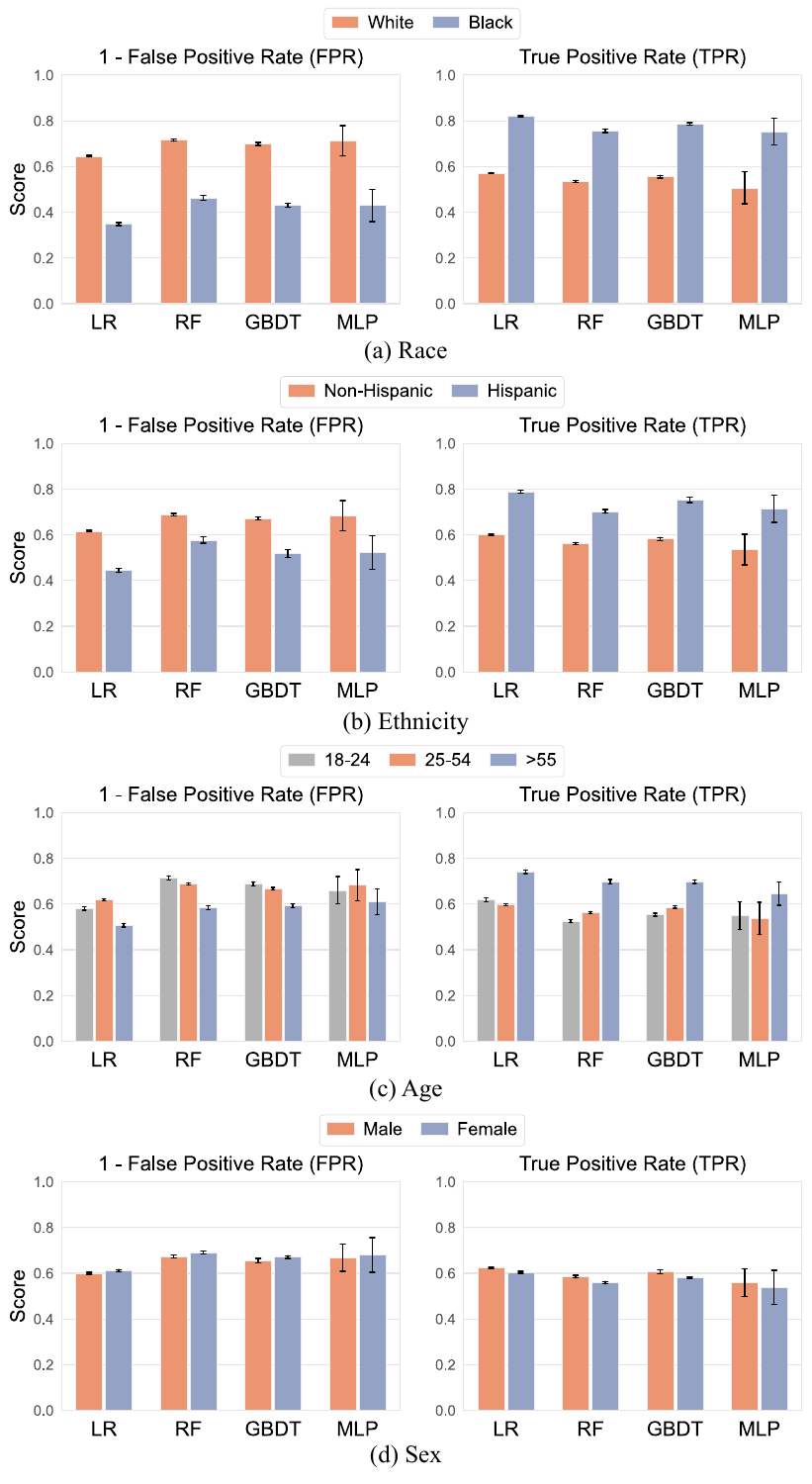}
\caption{FPRs and TPRs across different subgroups are evaluated for four ML models (LR, RF, GBDT, MLP) as the baseline models for predicting premature treatment discontinuation. The subfigures presents results stratified by (a) race, (b) ethnicity, (c) age, and (d) sex. Bars represent the mean values over 10 runs, while error bars indicate the standard deviation computed from these runs. Higher values indicate better results. }
\label{fig:dropout_baseline}
\end{figure}

\section{Results}

\subsection{Descriptive Results}
A total of 366,825 patients satisfied the respective inclusion/exclusion criteria for the model development. 
The cohort’s data characteristics are summarized in Table~\ref{tab:dataset}. 
Of all treatment episodes, 54.15\% resulted in premature discontinuation, while 38.65\% retained longer than 180 days. In terms of sociodemographic and clinical characteristics, the majority were aged 25–54 (79.63\%), White (84.14\%) and non-Hispanic patients (92.52\%). The majority of patients (72.9\%) had an education level of grade 12 or below, and 73.01\% were unemployed at the time of admission. Most patients (78.16\%) reported living in independent housing, and 55.44\% were from the Northeast region. Only 5.4\% reported having been arrested in the 30 days prior to admission. Additionally, 41.6\% had co-occurring psychological disorders, 54.6\% reported injection drug use, and 80.98\% had used heroin. 

\begin{table}[tbp]
\caption{Overall predictive performance, measured by area under the receiver operating characteristic curve (AUROC), for ML models predicting premature treatment discontinuation and treatment retention exceeding 180 days.}
\label{tab:overall_performance}
\begin{tabular*}{0.8\textwidth}{@{\extracolsep\fill}lcc}
\toprule 
& Premature Discontinuation & Retention $>$ 180 days \\
\midrule
LR & 0.6562 (0.0011) & 0.6559 (0.0011) \\
RF & 0.6863 (0.0013) & 0.6797 (0.0011) \\
GBDT & 0.6892 (0.0013) & 0.6836 (0.0010) \\
MLP & 0.6647 (0.0025) & 0.6657 (0.0014) \\
\bottomrule
\end{tabular*}
\end{table}

\begin{figure}[tbp]
\centering
\includegraphics[width=0.72\textwidth]{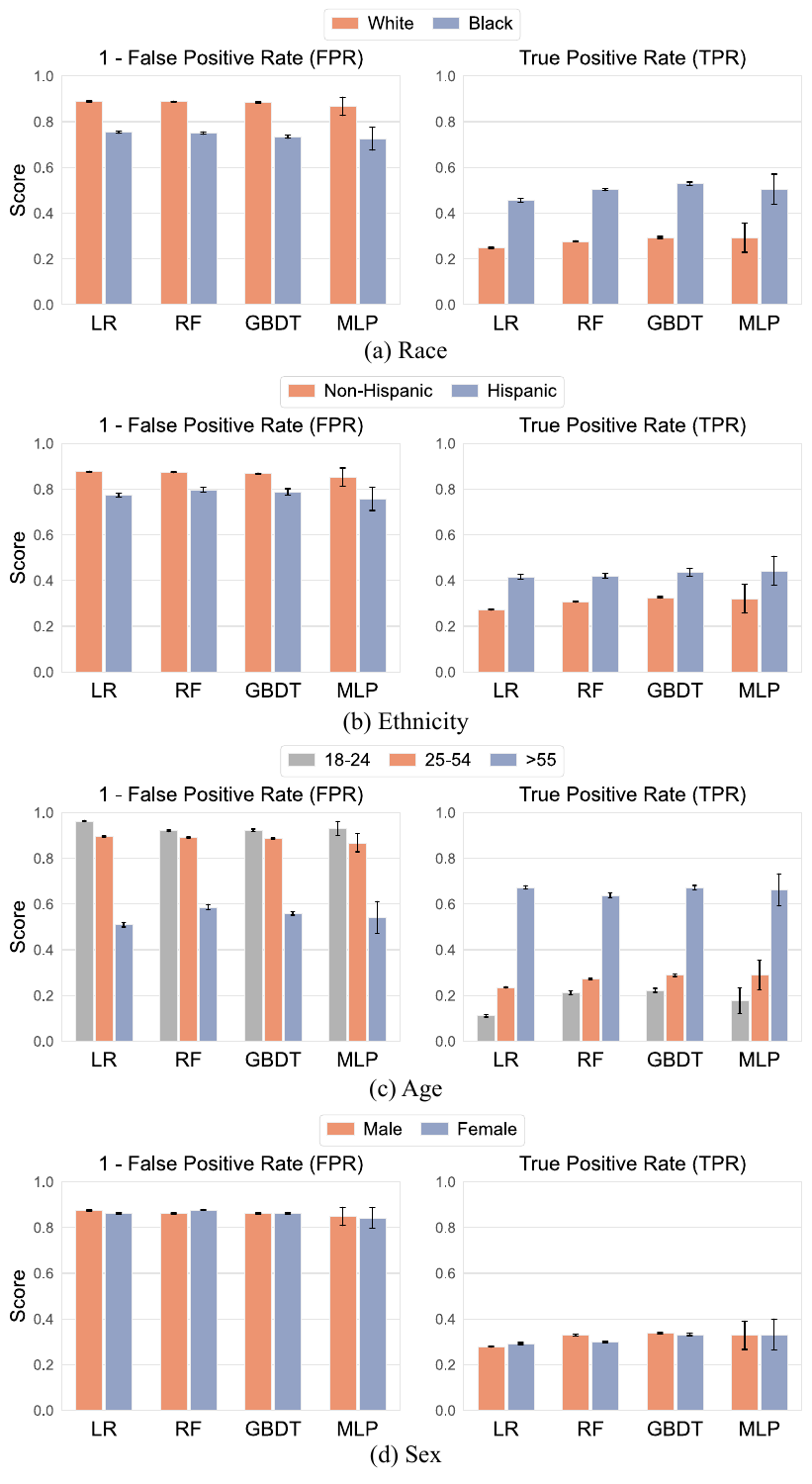}
\caption{FPRs and TPRs across different subgroups are evaluated for four ML models (LR, RF, GBDT, MLP) as the baseline models for predicting treatment retention $>$ 180 days. The subfigures presents results stratified by (a) race, (b) ethnicity, (c) age, and (d) sex. Bars represent the mean values over 10 runs, while error bars indicate the standard deviation computed from these runs. Higher values indicate better results. }
\label{fig:los_baseline}
\end{figure}

\subsection{Model Performance Across Patient Subgroups}
We began by systematically evaluating how each of the four ML models performed, with particular attention to how performance varied across patient subgroups for the two target outcomes. 
Table~\ref{tab:overall_performance} reports the overall AUROC values for predicting premature treatment discontinuation and treatment retention exceeding 180 days. Among the four models evaluated, the GBDT model achieved the highest AUROC. 
Figure~\ref{fig:dropout_baseline} presents subgroup-specific FPRs and TPRs for premature discontinuation predictions, stratified by different sensitive attributes for each of the four models. 
Across all models, FPRs were consistently higher for Black patients compared to White patients, with a gap of 29.67\% for LR, 25.26\% for RF, 26.81\% for GBDT, and 28.28\% for MLP, indicating a tendency to incorrectly classify Black patients as likely to drop out of treatment prematurely. 
Gaps were also observed in TPRs, with higher TPRs for Black patients across all models (LR: 24.91\%, RF: 22.13\%, GBDT: 23.04\%, MLP: 24.63\%). 
Our results show that the FPR and TPR gaps across ethnicity (11\%–19\%), age (7\%–18\%), and sex ($<$ 3\%) subgroups are less pronounced, compared to those observed for race. 

Figure~\ref{fig:los_baseline} shows FPRs and TPRs for predicting treatment retention $>$ 180 days. 
Our results indicate that both Black patients and patients over 55 years old consistently exhibited higher FPRs and TPRs compared to White patients and younger age subgroups, respectively. 
The FPR gap between Black patients and White patients ranged from 13\% to 15\% across all models, while TPR gaps were even larger, ranging from 20\% to 24\%. 
Age-related gaps were evident, with older patients ($>$ 55) having 33\%–46\% higher FPRs and 42\%–57\% higher TPRs than younger subgroups across all models. 
The gaps were smaller among ethnic subgroups (7\%–15\%) and even lower among sex subgroups ($<$ 3\%). 
For comparison, we also trained models excluding sensitive attributes as input features (see Tables \ref{tab:ablation}-\ref{tab:ablation2}). 
To provide a threshold-independent assessment of model performance across demographic groups, we additionally computed the AUROC separately for each subgroup defined by race, ethnicity, and age. The detailed results are presented in Appendix~\ref{subgruop_AUROC}. In addition, to evaluate the robustness of our results under post-pandemic changes in MOUD treatment, we additionally analyzed the 2020–2023 TEDS-D data using the same experimental pipeline. The results, presented in Appendix~\ref{results_20_23}, showed similar fairness patterns and slightly improved predictive performance, supporting the robustness of our findings across different time periods. 

\begin{figure}[tbp]
\centering
\includegraphics[width=0.8\textwidth]{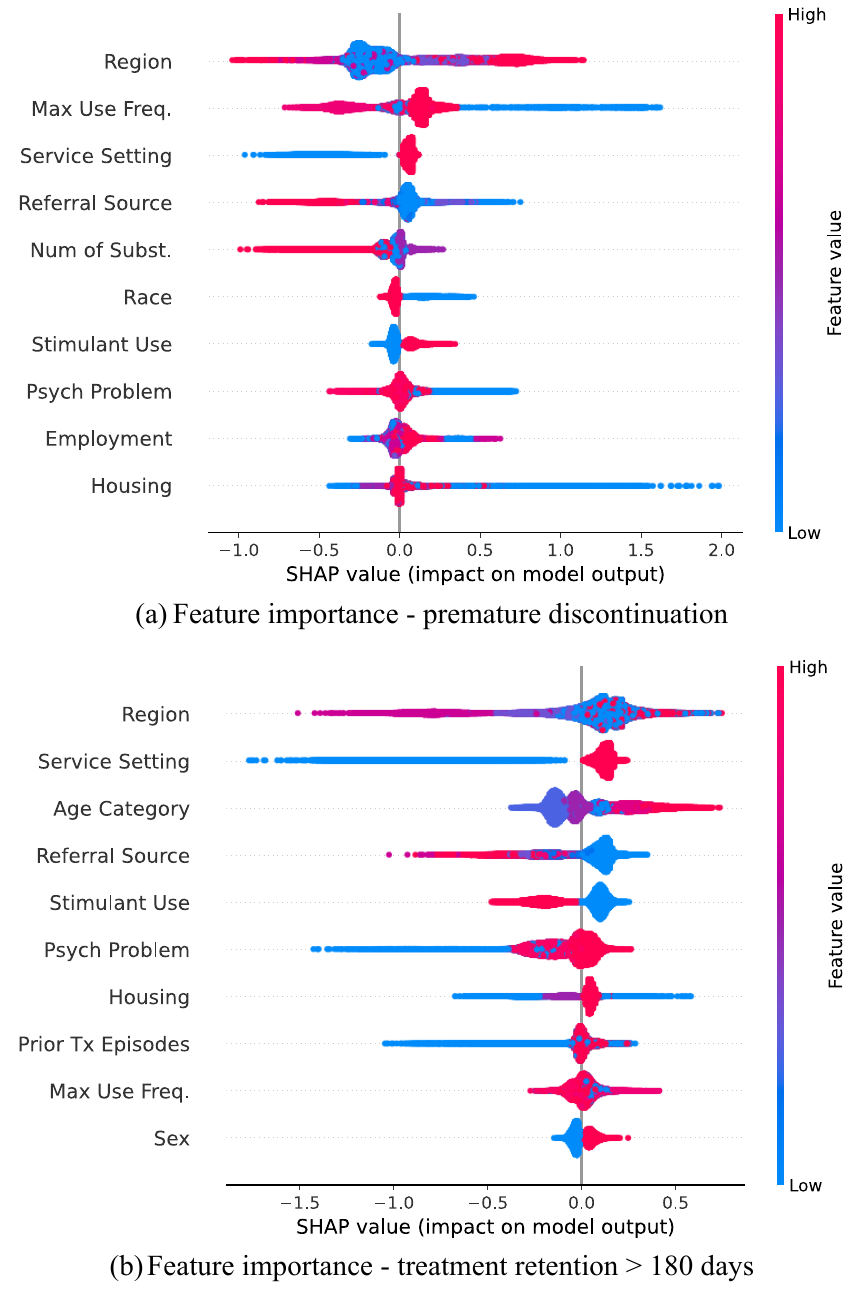}
\caption{Figures (a) and (b) present feature importance plots of the GBDT model for two prediction tasks: (a) premature treatment discontinuation and (b) treatment retention $>$ 180 days. The x-axis represents feature attribution values, where positive values indicate that a feature increases the likelihood of the positive class, while negative values indicate a decrease. Features are ranked by their average importance, with the most influential ones appearing at the top.} 
\label{fig:shap_dropout_GBDT}
\end{figure}

\begin{figure}[tbp]
\centering
\includegraphics[width=\textwidth]{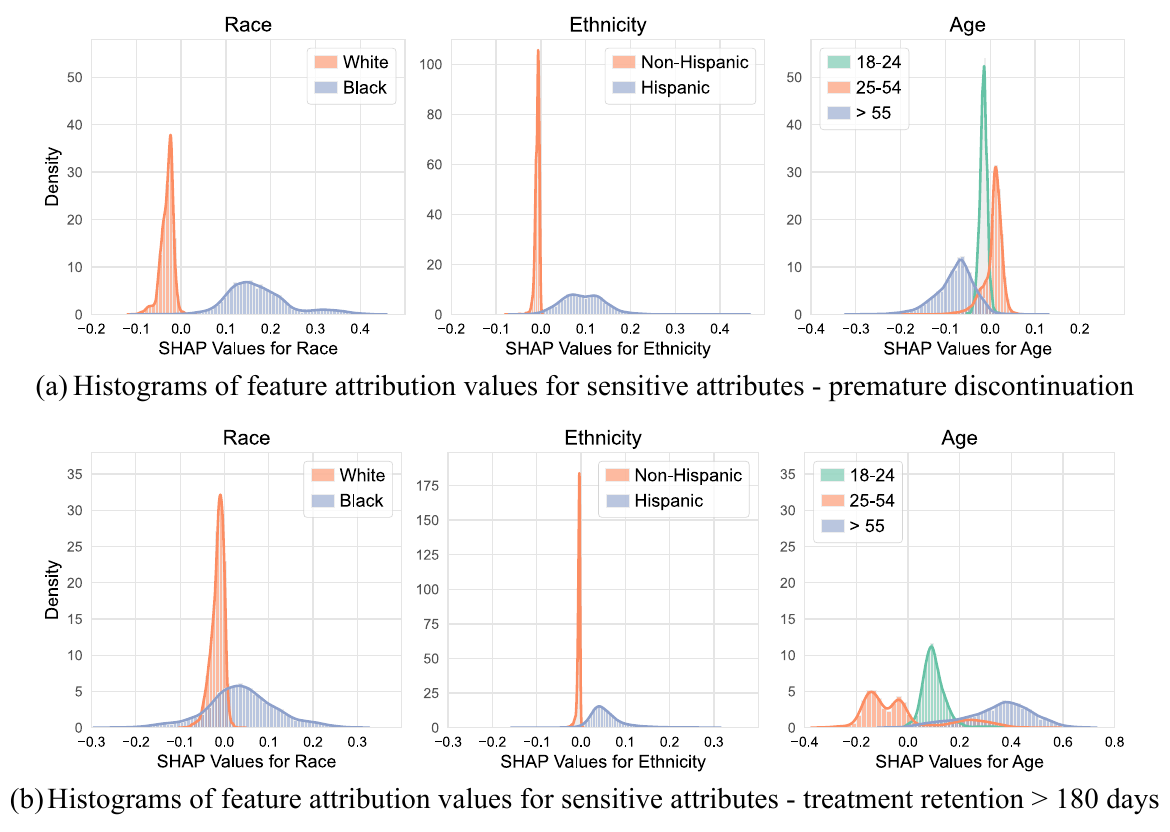}
\caption{Figures (a) and (b) display histograms of feature attribution values for sensitive attributes (race, ethnicity, and age) for each task: (a) premature treatment discontinuation and (b) treatment retention $>$ 180 days, to visualize how these attributes contribute to model predictions.} 
\label{fig:sensitive_attr_dropout_GBDT}
\end{figure}

\subsection{Interpreting Model Behavior Through Feature Attribution}
To interpret how input features influenced the model’s predictions, we computed SHAP values as feature attributions, focusing on the GBDT model, which demonstrated the best overall performance among the four models evaluated. 
In our analysis, all categorical predictors were one-hot encoded prior to model training, and SHAP values were initially computed at the level of the one-hot encoded features. To facilitate interpretation, we aggregated SHAP values back to the original categorical variables by summing the SHAP values of all one-hot encoded columns corresponding to each feature, following common practice. So, the reported SHAP values represent the overall contribution of each original categorical feature, reflecting the combined effect of all its encoded categories. 
Figure~\ref{fig:shap_dropout_GBDT} presents a global explanation of the GBDT model’s predictions for (a) premature treatment discontinuation and (b) treatment retention exceeding 180 days. It displays the top 10 most influential features, ranked by their mean absolute SHAP values aggregated over the test set. 
The feature attribution values indicate each feature’s contribution to increasing (positive SHAP values) or decreasing (negative SHAP values) the likelihood of positive class, while the color gradient represents the feature value. 
For both outcome variables, the features that consistently had strong influence on model's predictions included geographic region, patient's referral source, type of service setting (intensive or non-intensive outpatient), max frequency of substance use (i.e., heroin, non-prescription methadone, other opiates and synthetics), the presence of co-occurring mental or behavioral health disorders. 

As part of our analysis, we examined whether the observed performance differences reflected systematic variation in how the model treated different patient subgroups—that is, whether the model relied on sensitive attributes in a way that led to consistently different predictions across subgroups. 
Figure~\ref{fig:sensitive_attr_dropout_GBDT} shows the distributions of feature attributions for sensitive attributes over individual test points from different subgroups. In predicting premature treatment discontinuation, race demonstrated relatively high feature attribution values, indicating that the model relies heavily on this attribute, whereas age and ethnicity contributed less. 
Notably, the effect of race varied across subgroups: feature attribution values were predominantly positive for Black patients and negative for White patients, indicating differing influences on predicted risk. 
Similarly, for treatment retention prediction, age emerged as a key predictive factor, as reflected in its higher SHAP values. The impact of age also varied across subgroups, with patients aged over 55 receiving the most positive feature attribution values, suggesting the model associated older age with a higher likelihood of being retained in treatment for more than 180 days. 
Table~\ref{tab:white_black_comparison} compares the feature profiles of false positives for White and Black patients. 
Black patients who were falsely classified as at risk of premature discontinuation or as likely to remain in treatment beyond 180 days were generally older, had lower education levels, and exhibited higher rates of unemployment compared to the White patients. Additionally, heroin use was more prevalent among Black patients, while injection drug use was more common among White patients. 
These patterns suggest that the model may rely on different combinations of features—some of which may act as proxies for sensitive attributes—when making predictions for different subgroups. 

\begin{table}[tbp]
\caption{Sample characteristics of false positive subgroup from GBDT model for premature treatment discontinuation and retention $>$ 180 days, comparing Black and White patients.}
\label{tab:white_black_comparison}
\begin{tabular*}{\textwidth}{@{\extracolsep\fill}lcccc}
\toprule 
& \multicolumn{2}{@{}c@{}}{\textbf{Premature Discontinuation}} 
& \multicolumn{2}{@{}c@{}}{\textbf{Retention $>$ 180 days}} \\
\cmidrule(lr){2-3} \cmidrule(l){4-5} 

& \makecell{False Positive \\ White patients \\ (\%)} & \makecell{False Positive \\ Black patients \\ (\%)}  & \makecell{False Positive \\ White patients \\ (\%)} & \makecell{False Positive \\ Black patients \\ (\%)} \\
 
\midrule
Age  &  &  &  &  \\
\quad 18-24 & 11.3 & 2.8  & 7.1 & 0.5 \\
\quad 25-54 & 82.6 & 67.6  & 72.6 & 54.5 \\
\quad $>$ 55 & 6.1 & 29.6  & 20.3 & 45.0 \\
Education  &  &  & & \\
\quad Less than grade 11  & 23.9 & 35.2 & 26.2 & 33.7\\
\quad Grade 12 & 44.6 & 45.5 & 43.7 & 45.6 \\
\quad College or more & 30.0 & 18.3 & 26.7 & 19.7\\
Employment status  &  &  & &  \\
\quad Full-time/Part-time & 25.3 & 12.5 & 32.8 & 13.3 \\
\quad Unemployed & 73.4 & 83.7 & 64.8 & 86.0\\
Used Heroin & 80.7 & 91.4 & 79.3 & 94.0 \\
Used by injection & 59.2 & 28.2  & 53.3 & 27.5 \\

\bottomrule
\end{tabular*}
\end{table}

\begin{figure}[tbp]
\centering
 \includegraphics[width=0.98\textwidth]{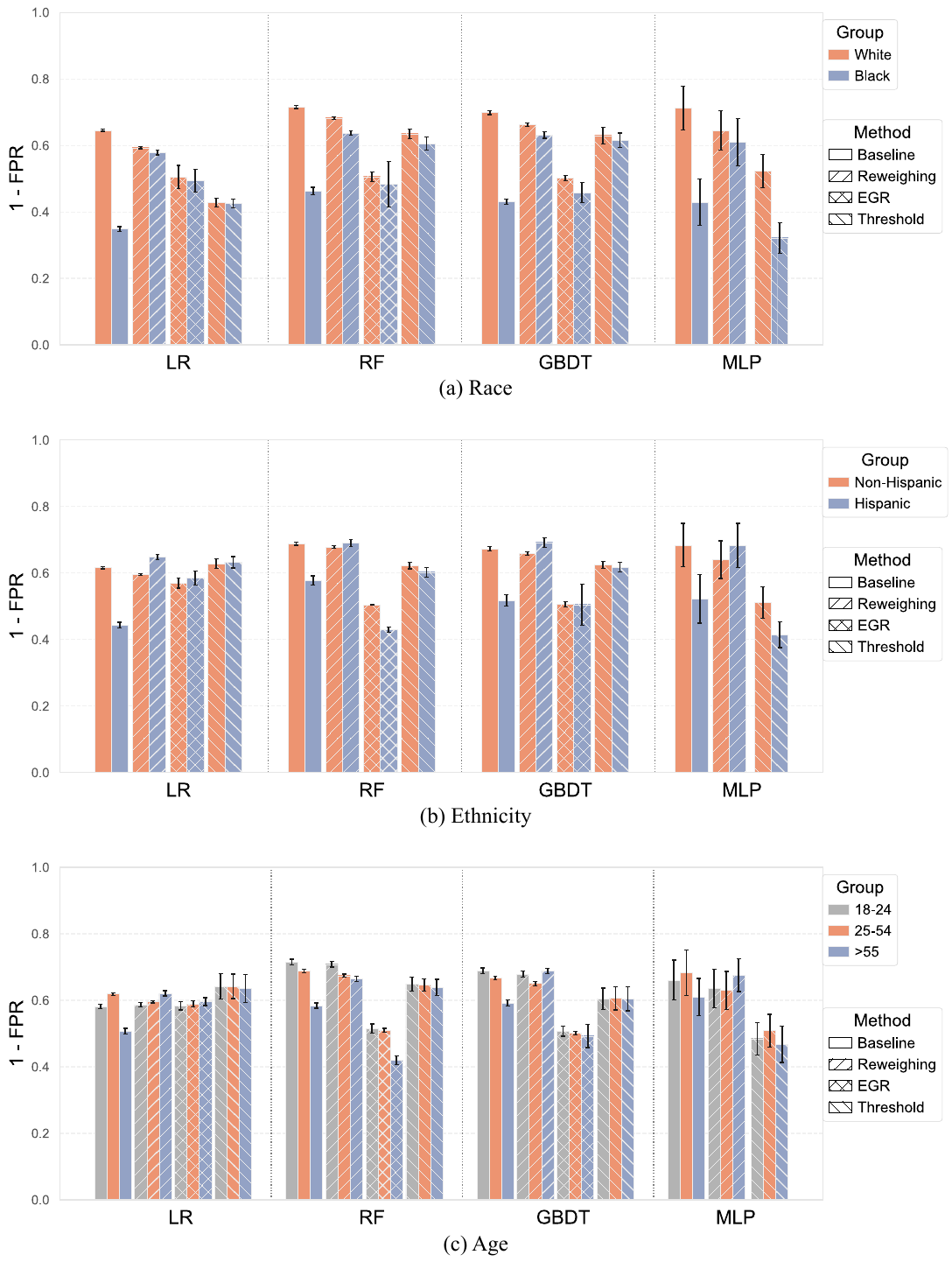}
\caption{FPRs across different subgroups are shown for three bias mitigation methods—Reweighing (pre-processing), EGR (in-processing), and Threshold (post-processing)—in predicting premature treatment discontinuation, compared to a baseline without fairness considerations. Results are stratified by (a) race, (b) ethnicity, and (c) age. Bars represent the mean values over 10 independent runs, with error bars indicating the standard deviation across these runs. Higher values indicate better results. } 
\label{fig:dropout_bias_mitigation_FPR}
\end{figure}

\begin{figure}[tbp]
\centering
 \includegraphics[width=0.98\textwidth]{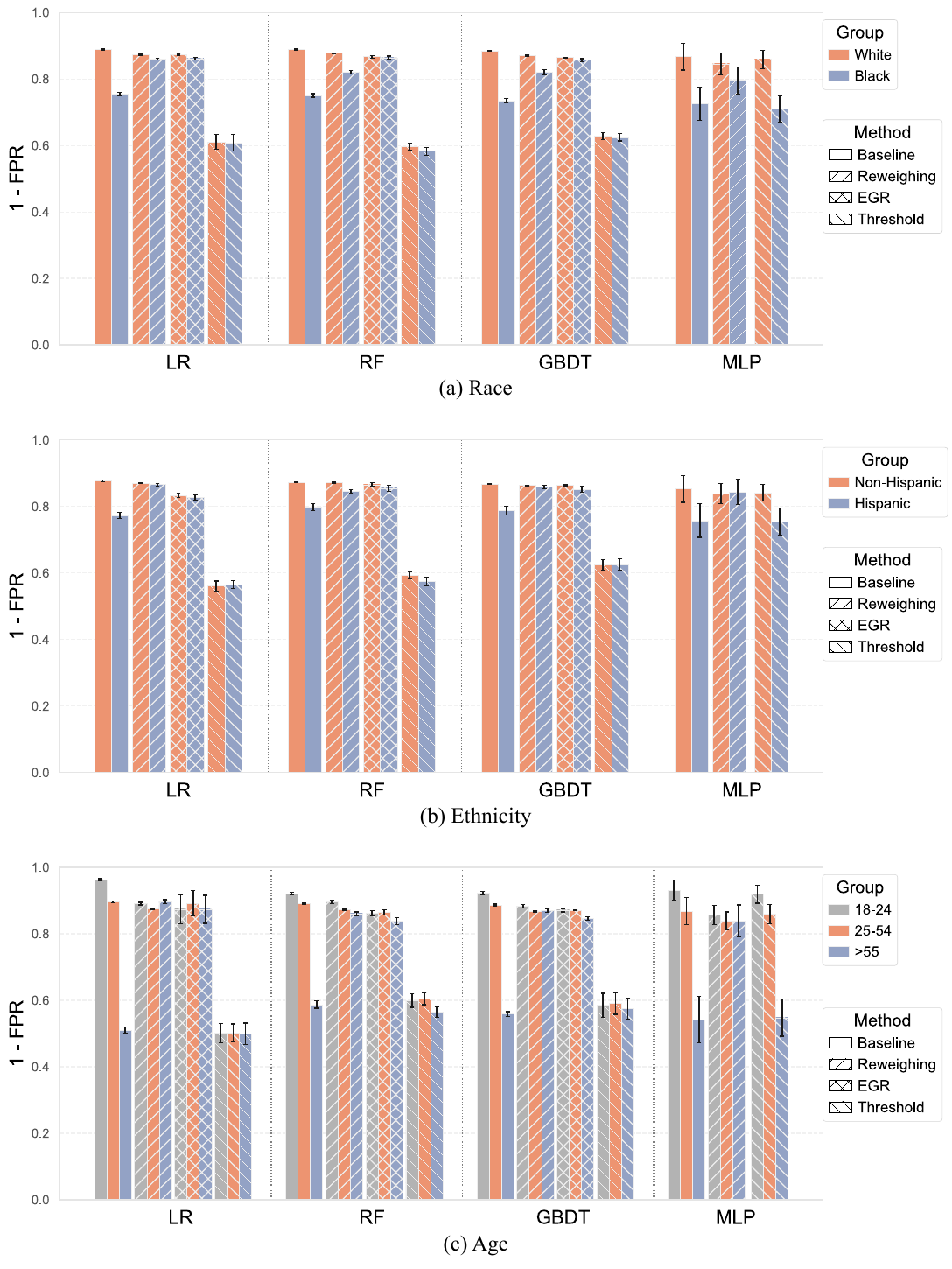}
\caption{FPRs across different subgroups are shown for three bias mitigation methods—Reweighing (pre-processing), EGR (in-processing), and Threshold (post-processing)—in predicting treatment retention $>$ 180 days, compared to a baseline without fairness considerations. Results are stratified by (a) race, (b) ethnicity, and (c) age. Bars represent the mean values over 10 independent runs, with error bars indicating the standard deviation across these runs. Higher values indicate better results.} 
\label{fig:los_bias_mitigation_FPR}
\end{figure}

\subsection{Bias Mitigation Results}
To address the observed performance inconsistencies across subgroups, we implemented and assessed three bias mitigation strategies, including Reweighing \cite{kamiran2012data}, EGR \cite{agarwal2018reductions}, and Threshold \cite{hardt2016equality}. 
As shown in Figure~\ref{fig:dropout_bias_mitigation_FPR}, for predicting premature discontinuation, Reweighing significantly reduced gaps in FPR (LR: 1.36\%, RF: 4.52\%, GBDT: 3.19\%, MLP: 4.34\%) between Black and White patients. 
Similarly, EGR reduced FPR gaps to 1.08\% for LR, 4.89\% for RF, and 4.49\% for GBDT. 
Note that EGR is not compatible with MLPs, as they are non-convex and typically do not support per-sample reweighting in a stable manner as required. 
Threshold reduced FPR gaps, achieving the lowest gaps in LR (0.56\%), RF (3.06\%), and GBDT (1.36\%) between White and Black patients, but was less effective for MLP (19.89\%). 
It is important to note that, although all three mitigation methods reduced subgroup gaps, these improvements often came at the expense of higher error rates for at least one subgroup. For example, the Threshold method narrowed the gap by worsening the FPR for White patients while improving it for Black patients, and in some cases, both subgroups experienced worsened FPRs even though the overall gap was reduced. 
For age subgroups, Reweighing reduced the FPR gap (LR: 2.63\%, RF: 0.43\%, GBDT: 7.54\%, MLP: 7.4\%) between younger (18-54) and older ($>$ 55) subgroups, compared to the baseline models (LR: 10.75\%, RF: 8.69\%, GBDT: 5.29\%, MLP: 4.39\%). EGR brings FPR gaps to 0.7\% (LR), 2.47\% (RF), and 1.0\% (GBDT), while Threshold produced the most minimal FPR gaps (LR: 0.71\%, RF: 1.24\%, GBDT: 0.15\%, MLP: 2.84\%). 
Similarly, for ethnicity, all three bias mitigation methods reduced the overall FPR gaps (Reweighing—LR: 5.22\%, RF: 1.16\%, GBDT: 3.33\%, MLP: 4.28\%; EGR—LR: 1.53\%, RF: 7.41\%, GBDT: 4.53\%; Threshold—LR: 0.99\%, RF: 2.03\%, GBDT: 0.88\%, MLP: 9.75\%). However, these reductions were often achieved by worsening the FPR for one subgroup or, in some cases, for both subgroups. 
Such patterns were consistently observed across all three mitigation methods and across the subgroups examined. 
A similar trend was observed in Figure~\ref{fig:los_bias_mitigation_FPR} for predicting treatment retention $>$ 180 days, where all three mitigation strategies led to reductions in FPR gaps across subgroups with varying effectiveness. 
Specifically, the FPR gap across race subgroups was substantially reduced by Reweighing (LR: 1.43\%, RF: 5.71\%, GBDT: 4.97\%, MLP: 5.00\%) and by EGR (LR: 1.21\%, RF: 0.37\%, GBDT: 0.78\%). The Threshold method also reduced race-based FPR gaps for several models (LR: 0.47\%, RF: 1.34\%, GBDT: 0.49\%); however, for the MLP model, threshold adjustment instead exacerbated the gap, increasing the FPR gap to 14.87\%. 
Similary, the FPR gap across age subgroups was significantly reduced with Reweighing (LR: 2.17\%, RF: 3.52\%, GBDT: 1.74\%, MLP: 3.55\%) and EGR (LR: 2.06\%, RF: 2.94\%, GBDT: 2.73\%). The Threshold method also substantially reduced the FPR gap for some models (LR: 0.96\%, RF: 4.01\%, GBDT: 1.70\%), but the gap persisted at 37.16\% for MLP. 
For ethnicity subgroups, although the baseline gaps were relatively modest, all three mitigation methods generally reduced FPR gaps compared to the Baseline (LR: 10.30\%, RF: 7.59\%, GBDT: 8.05\%, MLP: 9.52\%). Reweighing reduced the gaps to 0.50\% for LR, 2.60\% for RF, 0.58\% for GBDT, and 1.19\% for MLP. EGR reduced the gaps to 0.71\% for LR, 1.08\% for RF, and 1.23\% for GBDT. The Threshold method also lowered the gaps for several models (LR: 0.90\%, RF: 1.82\%, GBDT: 0.97\%), but resulted in a large gap for MLP (8.65\%). 
All TPR results are presented in Figures \ref{fig:dropout_bias_mitigation_TPR} and \ref{fig:los_bias_mitigation_TPR}, with more detailed results in Tables \ref{tab:all_FPR_dropout}-\ref{tab:all_TPR_los}.  

\begin{sidewaystable}[tbp]
\caption{Group fairness metrics (mean; standard deviation in parentheses) for predicting premature treatment discontinuation. Predictive Equality Ratio: FPR ratio of the subgroup with the lowest FPR to the subgroup with the highest FPR. Equal Opportunity Ratio: TPR ratio of the subgroup with the lowest TPR to the subgroup with the highest TPR. Values closer to 1 indicate more equitable model performance across subgroups. \textbf{Bold} values indicate the best results.}
\label{tab:group_fairness_dropout}
\begin{tabular*}{\textheight}{@{\extracolsep\fill}llcccccccc@{}}
\toprule
\multirow{2}{*}{\textbf{Group}} & \multirow{2}{*}{\textbf{Method}} 
& \multicolumn{4}{c}{\textbf{Predictive Equality Ratio}} 
& \multicolumn{4}{c}{\textbf{Equal Opportunity Ratio}} \\

\cmidrule(lr){3-6} \cmidrule(l){7-10} 
&  & LR & RF & GBDT & MLP & LR & RF & GBDT & MLP \\

\midrule
\multirow{8}{*}{Race} 
& Baseline 
& \makecell{0.5444 \\ (0.0054)} & \makecell{0.5296 \\ (0.0117)} & \makecell{0.5291 \\ (0.0123)} & \makecell{0.5010 \\ (0.0702)}    
& \makecell{0.6966 \\ (0.0034)} & \makecell{0.7075 \\ (0.0080)} & \makecell{0.7070 \\ (0.0089)} & \makecell{0.6715 \\ (0.0570)}    \\
\cmidrule(lr){3-6} \cmidrule(l){7-10} 
& Reweighing   
& \makecell{0.9679 \\ (0.0150)} & \makecell{0.8756 \\ (0.0099)} & \makecell{0.9141 \\ (0.0175)} & \textbf{\makecell{0.8941 \\ (0.0539)}}    
& \makecell{0.9537 \\ (0.0111)} & \makecell{0.9566 \\ (0.0125)} & \makecell{0.9802 \\ (0.0129)} & \textbf{\makecell{0.9509 \\ (0.0322)}}    \\
\cmidrule(lr){3-6} \cmidrule(l){7-10} 
& EGR    
& \makecell{0.9787 \\ (0.0147)} & \makecell{0.9104 \\ (0.0715)} & \makecell{0.9198 \\ (0.0522)} & -- \footnotemark[5]  
& \makecell{0.9849 \\ (0.0091)} & \makecell{0.9485 \\ (0.0505)} & \makecell{0.9659 \\ (0.0155)} & -- \footnotemark[5]    \\
\cmidrule(lr){3-6} \cmidrule(l){7-10} 
& Threshold    
& \textbf{\makecell{0.9888 \\ (0.0060)}} & \textbf{\makecell{0.9256 \\ (0.0126)}} & \textbf{\makecell{0.9645 \\ (0.0154)}} & \makecell{0.7024 \\ (0.0368)}    
& \textbf{\makecell{0.9959 \\ (0.0031)}} & \textbf{\makecell{0.9805 \\ (0.0080)}} & \textbf{\makecell{0.9896 \\ (0.0072)}} & \makecell{0.8344 \\ (0.0301)}    \\

\midrule
\multirow{8}{*}{Ethnicity} 
& Baseline     
& \makecell{0.6900 \\ (0.0120)} & \makecell{0.7382 \\ (0.0205)} & \makecell{0.6803 \\ (0.0314)} & \makecell{0.6578 \\ (0.0431)} 
& \makecell{0.7600 \\ (0.0081)} & \makecell{0.8000 \\ (0.0109)} & \makecell{0.7733 \\ (0.0194)} & \makecell{0.7479 \\ (0.0394)}   \\
\cmidrule(lr){3-6} \cmidrule(l){7-10} 
& Reweighing   
& \makecell{0.8710 \\ (0.0181)} & \textbf{\makecell{0.9638 \\ (0.0236)}} & \makecell{0.9023 \\ (0.0386)} & \textbf{\makecell{0.8753 \\ (0.0669)}}    
& \makecell{0.9739 \\ (0.0094)} & \makecell{0.9583 \\ (0.0116)} & \textbf{\makecell{0.9823 \\ 0.0153}} & \textbf{\makecell{0.9677 \\ (0.0292)}}    \\
\cmidrule(lr){3-6} \cmidrule(l){7-10} 
& EGR      
& \makecell{0.9640 \\ (0.0208)} & \makecell{0.8701 \\ (0.0119)} & \makecell{0.9127 \\ (0.0592)} & --    
& \makecell{0.9871 \\ (0.0101)} & \makecell{0.8991 \\ (0.0066)} & \makecell{0.9510 \\ (0.0307)} & --    \\
\cmidrule(lr){3-6} \cmidrule(l){7-10} 
& Threshold    
& \textbf{\makecell{0.9737 \\ (0.0220)}} & \makecell{0.9495 \\ (0.0193)} & \textbf{\makecell{0.9776 \\ (0.0230)}} & \makecell{0.8325 \\ (0.0359)}    
& \textbf{\makecell{0.9894 \\ (0.0088)}} & \textbf{\makecell{0.9906 \\ (0.0065)}} & \makecell{0.9903 \\ (0.0063)} & \makecell{0.8851 \\ (0.0259)}    \\

\midrule
\multirow{8}{*}{Age} 
& Baseline    
& \makecell{0.7734 \\ (0.0171)} & \makecell{0.6857 \\ (0.0214)} & \makecell{0.7644 \\ (0.0213)} & \makecell{0.8061 \\ (0.0806)} 
& \makecell{0.8056 \\ (0.0117)} & \makecell{0.7527 \\ (0.0157)} & \makecell{0.7932 \\ (0.0120)} & \makecell{0.8219 \\ (0.0526)}    \\
\cmidrule(lr){3-6} \cmidrule(l){7-10} 
& Reweighing   
& \makecell{0.9167 \\ (0.0249)} & \makecell{0.8674 \\ (0.0250)} & \makecell{0.8889 \\ (0.0258)} & \makecell{0.8727 \\ (0.0655)}    
& \makecell{0.9699 \\ (0.0106)} & \makecell{0.8511 \\ (0.0138)} & \makecell{0.9253 \\ (0.0141)} & \textbf{\makecell{0.9481 \\ (0.0310)}}    \\
\cmidrule(lr){3-6} \cmidrule(l){7-10} 
& EGR          
& \makecell{0.9590 \\ (0.0141)} & \makecell{0.8259 \\ (0.0248)} & \makecell{0.9225 \\ (0.0384)} & --    
& \makecell{0.9713 \\ (0.0119)} & \makecell{0.8747 \\ (0.0147)} & \makecell{0.9421 \\ (0.0430)} & --   \\
\cmidrule(lr){3-6} \cmidrule(l){7-10} 
& Threshold  
& \textbf{\makecell{0.9706 \\ (0.0147)}} & \textbf{\makecell{0.9628 \\ (0.0169)}} & \textbf{\makecell{0.9713 \\ (0.0099)}} & \textbf{\makecell{0.9167 \\ (0.0300)}}    
& \textbf{\makecell{0.9806 \\ (0.0060)}} & \textbf{\makecell{0.9790 \\ (0.0106)}} & \textbf{\makecell{0.9820 \\ (0.0081)}} & \makecell{0.9259 \\ (0.0169)}    \\

\bottomrule
\end{tabular*}
\footnotetext[5]{Results for MLP with EGR are not reported, as EGR requires models that support stable, per-sample reweighting during training. MLP are non-convex and typically do not support per-sample reweighting in a stable and standardized manner as required.} 
\end{sidewaystable}

\begin{sidewaystable}[tbp]
\caption{Group fairness metrics (mean; standard deviation in parentheses) for predicting treatment retention $>$ 180 days. Predictive Equality Ratio: FPR ratio of the subgroup with the lowest FPR to the subgroup with the highest FPR. Equal Opportunity Ratio: TPR ratio of the subgroup with the lowest TPR to the subgroup with the highest TPR. Values closer to 1 indicate more equitable model performance across subgroups. \textbf{Bold} values indicate the best results.}
\label{tab:group_fairness_los}
\begin{tabular*}{\textheight}{@{\extracolsep\fill}llcccccccc@{}}
\toprule
\multirow{2}{*}{\textbf{Group}} & \multirow{2}{*}{\textbf{Method}} 
& \multicolumn{4}{c}{\textbf{Predictive Equality Ratio}} 
& \multicolumn{4}{c}{\textbf{Equal Opportunity Ratio}} \\

\cmidrule(lr){3-6} \cmidrule(l){7-10} 
&  & LR & RF & GBDT & MLP & LR & RF & GBDT & MLP \\

\midrule
\multirow{8}{*}{Race} 
& Baseline     
& \makecell{0.4501 \\ (0.0108)} & \makecell{0.4468 \\ (0.0113)} & \makecell{0.4351 \\ (0.0133)} & \makecell{0.4803 \\ (0.0823)} 
& \makecell{0.5440 \\ (0.0091)} & \makecell{0.5514 \\ (0.0069)} & \makecell{0.5554 \\ (0.0096)} & \makecell{0.5775 \\ (0.0769)}    \\
\cmidrule(lr){3-6} \cmidrule(l){7-10} 
& Reweighing   
& \makecell{0.8984 \\ (0.0164)} & \makecell{0.6828 \\ (0.0213)} & \makecell{0.7241 \\ (0.0355)} & \textbf{\makecell{0.7659 \\ (0.1286)}} 
& \makecell{0.9490 \\ (0.0143)} & \makecell{0.7503 \\ (0.0107)} & \makecell{0.8007 \\ (0.0179)} & \textbf{\makecell{0.8109 \\ (0.0943)}}    \\
\cmidrule(lr){3-6} \cmidrule(l){7-10} 
& EGR          
& \makecell{0.9129 \\ (0.0161)} & \textbf{\makecell{0.9732 \\ (0.0377)}} & \makecell{0.9470 \\ (0.0351)} & -- \footnotemark[5]   
& \makecell{0.9625 \\ (0.0225)} & \makecell{0.9747 \\ (0.0238)} & \makecell{0.9771 \\ (0.0176)} & -- \footnotemark[5]   \\
\cmidrule(lr){3-6} \cmidrule(l){7-10} 
& Threshold   
& \textbf{\makecell{0.9882 \\ (0.0089)}} & \makecell{0.9679 \\ (0.0121)} & \textbf{\makecell{0.9869 \\ (0.0105)}} & \makecell{0.4863 \\ (0.0491)} 
& \textbf{\makecell{0.9898 \\ (0.0093)}} & \textbf{\makecell{0.9862 \\ (0.0083)}} & \textbf{\makecell{0.9865 \\ (0.0068)}} & \makecell{0.5866 \\ (0.0431)}    \\

\midrule
\multirow{8}{*}{Ethnicity} 
& Baseline     
& \makecell{0.5457 \\ (0.0224)} & \makecell{0.6262 \\ (0.0318)} & \makecell{0.6234 \\ (0.0396)} & \makecell{0.6019 \\ (0.0614)} 
& \makecell{0.6552 \\ (0.0154)} & \makecell{0.7309 \\ (0.0160)} & \makecell{0.7485 \\ (0.0259)} & \makecell{0.7200 \\ (0.0608)}    \\
\cmidrule(lr){3-6} \cmidrule(l){7-10} 
& Reweighing   
& \makecell{0.9635 \\ (0.0246)} & \makecell{0.8326 \\ (0.0326)} & \makecell{0.9602 \\ (0.0290)} & \textbf{\makecell{0.9211 \\ (0.0582)}} 
& \makecell{0.9705 \\ (0.0114)} & \makecell{0.8918 \\ (0.0186)} & \makecell{0.9753 \\ (0.0153)} & \textbf{\makecell{0.9278 \\ (0.0534)}}    \\
\cmidrule(lr){3-6} \cmidrule(l){7-10} 
& EGR          
& \makecell{0.9595 \\ (0.0272)} & \makecell{0.9270 \\ (0.0424)} & \makecell{0.9194 \\ (0.0450)} & --    
& \makecell{0.9640 \\ (0.0220)} & \makecell{0.9656 \\ (0.0229)} & \makecell{0.9772 \\ (0.0178)} & --    \\
\cmidrule(lr){3-6} \cmidrule(l){7-10} 
& Threshold    
& \textbf{\makecell{0.9897 \\ (0.0100)}} & \textbf{\makecell{0.9577 \\ (0.0234)}} & \textbf{\makecell{0.9745 \\ (0.0161)}} & \makecell{0.6498 \\ (0.0496)} 
& \textbf{\makecell{0.9862 \\ (0.0112)}} & \textbf{\makecell{0.9794 \\ (0.0112)}} & \textbf{\makecell{0.9825 \\ (0.0109)}} & \makecell{0.7658 \\ (0.0420)}    \\

\midrule
\multirow{8}{*}{Age} 
& Baseline     
& \makecell{0.0755 \\ (0.0049)} & \makecell{0.1922 \\ (0.0107)} & \makecell{0.1744 \\ (0.0113)} & \makecell{0.1473 \\ (0.0496)} 
& \makecell{0.1650 \\ (0.0088)} & \makecell{0.3305 \\ (0.0154)} & \makecell{0.3313 \\ (0.0161)} & \makecell{0.2650 \\ (0.0641)}    \\
\cmidrule(lr){3-6} \cmidrule(l){7-10} 
& Reweighing   
& \makecell{0.8255 \\ (0.0359)} & \makecell{0.7479 \\ (0.0356)} & \makecell{0.8708 \\ (0.0407)} & \textbf{\makecell{0.7917 \\ (0.0710)}} 
& \makecell{0.6824 \\ (0.0226)} & \makecell{0.8373 \\ (0.0189)} & \makecell{0.8608 \\ (0.0316)} & \textbf{\makecell{0.8211 \\ (0.1003)}}    \\
\cmidrule(lr){3-6} \cmidrule(l){7-10} 
& EGR         
& \makecell{0.8358 \\ (0.0514)} & \makecell{0.8199 \\ (0.0336)} & \makecell{0.8236 \\ (0.0324)} & --    
& \makecell{0.8925 \\ (0.0401)} & \makecell{0.9195 \\ (0.0398)} & \makecell{0.9579 \\ (0.0207)} & --    \\
\cmidrule(lr){3-6} \cmidrule(l){7-10} 
& Threshold   
& \textbf{\makecell{0.9813 \\ (0.0150)}} & \textbf{\makecell{0.9078 \\ (0.0149)}} & \textbf{\makecell{0.9596 \\ (0.0158)}} & \makecell{0.1756 \\ (0.0399)} 
& \textbf{\makecell{0.9843 \\ (0.0070)}} & \textbf{\makecell{0.9849 \\ (0.0087)}} & \textbf{\makecell{0.9859 \\ (0.0065)}} & \makecell{0.3211 \\ (0.0474)}    \\

\bottomrule
\end{tabular*}
\footnotetext[5]{Results for MLP with EGR are not reported, as EGR requires models that support stable, per-sample reweighting during training. MLP are non-convex and typically do not support per-sample reweighting in a stable and standardized manner as required.} 
\end{sidewaystable}

\begin{figure}[tbp]
\centering
\includegraphics[width=\textwidth]{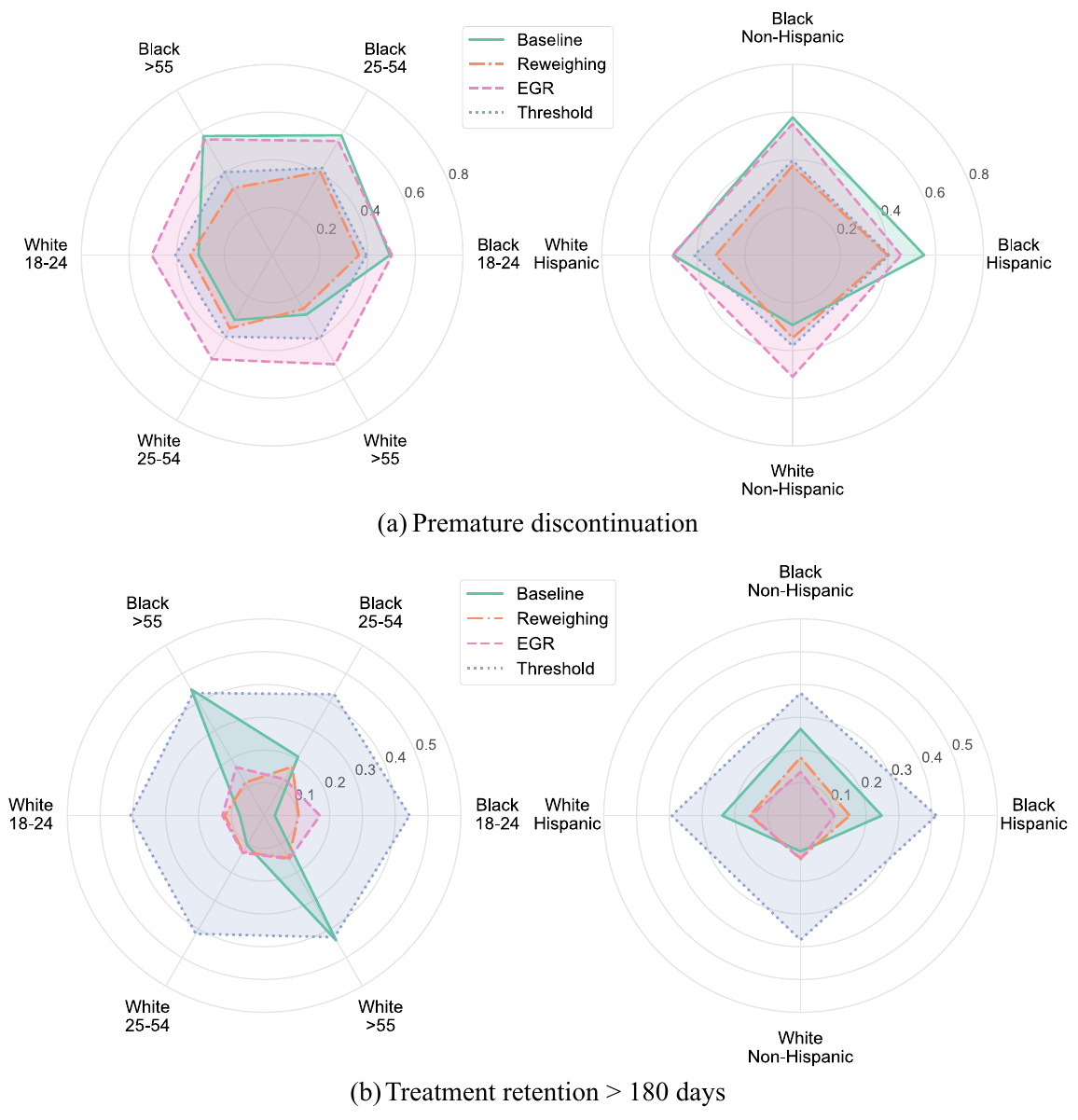}
\caption{FPRs for cross-tabulated subgroups across all bias mitigation methods in predicting two tasks (a) premature discontinuation and (b) treatment retention $>$ 180 days with GBDT model. Lower FPR values indicate better results.}
\label{fig:radar}
\end{figure}

We also calculated group fairness metrics, including EOR and PER, for each experiment to evaluate the equity of model performance across subgroups, as presented in Table~\ref{tab:group_fairness_dropout} and Table~\ref{tab:group_fairness_los}. 
All three bias mitigation methods enhanced both fairness metrics, bringing them closer to 1 compared to the baseline models. Among them, Threshold demonstrated significant fairness improvements in most experiments. 
However, it was less effective for the MLP model, likely because MLP models tend to produce more polarized probability outputs (i.e., predictions concentrated closer to 0 or 1), which leaves less flexibility for threshold adjustments to meaningfully alter subgroup-specific error rates. 
Moreover, the Threshold method often reduced subgroup gaps by worsening FPRs and TPRs for all the compared subgroups, as shown in the detailed subgroup results in Figures~\ref{fig:dropout_bias_mitigation_FPR}–\ref{fig:los_bias_mitigation_FPR} and Figures~\ref{fig:dropout_bias_mitigation_TPR}–\ref{fig:los_bias_mitigation_TPR}. 
Reweighing and EGR generally performed more consistently across the four models in improving fairness metrics. Notably, when examined alongside the subgroup-specific results, a trade-off between fairness and error rates remained evident across all three mitigation methods. 
In addition to EOR and PER, we also evaluated EOD to provide a unified assessment of gaps in both TPRs and FPRs. The detailed EOD results are provided in Table~\ref{tab:equalized_odds} in Appendix~\ref{EOD_results}. 

Finally, we conducted a comprehensive cross-tabulated subgroup evaluation using the best-performing GBDT model on both tasks to assess the effectiveness of bias mitigation strategies on more complex subgroups. These subgroups, defined by combinations of sensitive attributes such as race-age and race-ethnicity, allow us to explore more complex patterns in model behavior that may not be evident when attributes are assessed independently. 
Figure~\ref{fig:radar} presents the FPR across all cross-tabulated subgroup pairs for each mitigation strategy. Overall, we observe that the evaluated methods reduce average FPR and narrow the performance gaps across intersecting subgroups. This shows that these methods not only improve fairness along single dimensions, but also extend effectively to more complex, real-world group intersections.

\section{Discussion}
This study developed ML models and conducted a systematic assessment of their algorithmic fairness to predict MOUD treatment outcomes, including treatment retention and premature discontinuation, specifically in outpatient treatment settings. This setting represents a critical yet underexplored context in the MOUD literature, given the importance of sustained engagement for effective outcomes and the persistently high rates of premature discontinuation \cite{morgan2018injectable}. In outpatient MOUD care, treatment retention is shaped by a complex interplay of clinical, behavioral, and structural factors, and prior research has documented substantial differences in retention and completion across patient subgroups \cite{stahler2021racial,askari2020medication,baird2023determinants}. This study highlights the importance of evaluating not only overall predictive performance but also subgroup-level fairness when developing ML models in such settings. 
Across multiple models and outcome definitions, we observed consistent discrepancies in model performance across patient subgroups defined by sensitive attributes such as race, ethnicity, and age. Our findings demonstrate that models with similar overall predictive performance can nonetheless produce substantially different error rates across patient sociodemographic subgroups. In particular, we observed worse FPR for some patient subgroups, including Black patients and older patients, suggesting that the models may be relying on spurious patterns or proxies when making predictions. This observation aligns with prior findings showing that ML models often learn spurious correlations or non-generalizable patterns from training data, especially when model training data reflects long-standing patterns or encodes existing structural variation across patient subgroups \cite{yang2024identifying}. In the context of outpatient MOUD treatment, this is particularly concerning because outcome predictions may directly inform the allocation of evidence-based follow-up strategies known to improve typically low treatment retention, such as proactive appointment reminders and check-ins, early engagement with counseling or behavioral therapy, transportation support or telehealth adjustments, and peer recovery coaching \cite{warren2022using}. These tools may inadvertently amplify existing disparities in care and support if fairness is not explicitly examined. 

Furthermore, the prominent contributions of geographic region, patient referral source, service setting, substance use frequency, and co-occurring mental or behavioral health disorders to model predictions are consistent with prior studies identifying these factors as important determinants of MOUD treatment outcomes, including treatment retention and premature discontinuation \cite{stafford2022identifying,baird2023determinants}. 
Sensitive attributes such as race and age emerged as influential features in model predictions, but their impact varied across subgroups. 
Race contributed strongly to predictions of premature discontinuation, while age played a significant role in predicting long-term treatment retention. Notably, the impact of these features differed across subgroups, with being identified as Black individuals or being over 55 years old was associated with positive feature contributions in their respective tasks, while other subgroup characteristics showed negative or weaker associations. 
This pattern suggests that the model may consistently assign higher risk scores for Black patients for premature discontinuation and older patients with a greater likelihood of long-term retention. 
This raises concerns about whether the model is over-relying on subgroup-specific patterns in ways that could affect prediction reliability or lead to differential treatment recommendations. 
To further investigate this, we trained models without including sensitive attributes as input features (Tables \ref{tab:ablation}-\ref{tab:ablation2}). The overall prediction performance remained largely unchanged, suggesting that these attributes were not essential for predictive accuracy. 
This finding indicates that the models may have relied on such attributes as shortcuts for prediction, reflecting structural patterns present in the training data rather than identifying clinically relevant relationships. 
Such shortcut learning has been widely observed in ML systems, where models capture correlations that are statistically strong but clinically irrelevant or contextually misleading, as shown in previous studies \cite{mehrabi2021survey,weerts2024unlawful}. 
Notably, removing the sensitive attributes from the input features did not eliminate the observed performance discrepancies across subgroups, indicating that other correlated features may act as proxies. This finding underscores the difficulty of mitigating performance inconsistency simply by excluding subgroup identifiers alone, as structural signals may still be encoded through other variables. 


Our subgroup analysis of false positives revealed notable differences in the feature profiles of misclassified patients across subgroups. Specifically, Black patients who were incorrectly classified as at risk of premature discontinuation or as likely to remain in treatment long-term were generally older, less educated, more likely to be unemployed, and more likely to report heroin use. In contrast, misclassified White patients were more often associated with injection drug use. 
These findings echo long-standing evidence in public health literature showing that social determinants—such as education, employment, and access to stable resources—are deeply connected to health outcomes and treatment engagement \cite{obermeyer2019dissecting,bailey2017structural}. 
The prevalence of such social determinants among falsely flagged Black patients suggests that the model may be learning structural patterns from the training data, rather than focusing on clinical or behavioral risk factors. 
Furthermore, these trends indicate that false positive predictions may disproportionately affect patients with certain social determinants, reflecting how model predictions are influenced by such non-clinical factors. 
Although such feature dependencies may reflect real-world associations in the training data, they also raise important concerns about whether models are indirectly using sensitive characteristics through correlated variables. These findings highlight the need for scrutinizing which features drive model predictions to understand how models make decisions and to identify when outputs may reflect non-clinical factors embedded in data. 

To address these concerns, we implemented mitigation techniques aimed at reducing performance gaps across patient population subgroups. These methods targeted different stages of the ML development pipeline, including pre-processing, in-processing, and post-processing. While these approaches could improve subgroup-level fairness metrics, such as equal opportunity and predictive equality, they also introduced important trade-offs that require careful consideration. 
For example, reducing the FPR for one subgroup sometimes increased it for another, potentially leading to unintended consequences such as misallocated clinical resources or inappropriate interventions. 
These findings highlight that fairness interventions must be assessed not only through the lens of statistical parity, but also with attention to their real-world clinical and operational implications. A technically fair model may still produce disproportionate harm if misclassification carries different consequences for different subgroups. 
As such, healthcare providers and clinicians should scrutinize model outputs rather than treating them as neutral tools. From a policy perspective, these findings suggest the need for formal guidelines and validation standards that ensure predictive fairness. The demonstrated effectiveness of mitigation techniques further supports the integration of routine subgroup audits and fairness-aware practices into health system development processes. 

On the other hand, no single bias mitigation strategy serves as a silver bullet, and the choice of method should be guided by real-world implementation constraints and organizational capacity. 
Pre-processing methods such as Reweighing are often useful when models can be retrained but a low-friction approach is preferred that does not modify the learning algorithm itself \cite{kamiran2012data,chinta2025ai}. In practice, because Reweighing operates entirely before model fitting by adjusting sample weights, it can be incorporated into standard training workflows with minimal engineering overhead and is often a reasonable first-line option for mitigating disparities arising from imbalanced representation in data. 
In-processing methods such as EGR are best suited for scenarios in which organizations can retrain models and seek explicit control over fairness constraints during training \cite{agarwal2018reductions}. EGR is particularly appropriate when institutions want to enforce specific parity criteria (e.g., bounds on differences in TPR or FPR across groups) as part of the optimization objective and have sufficient technical capacity to integrate constraint-based training into their modeling pipelines. This approach offers a principled way to balance predictive performance and fairness during model development, rather than relying on post hoc adjustments. 
Post-processing methods such as the Threshold Optimizer are often appropriate when the underlying prediction model is fixed, for example, when using a vendor-supplied model or a model that has been operationally validated and cannot be easily retrained, and the primary degree of freedom lies in how predicted risks are translated into decisions. This is practically meaningful because many clinical and operational workflows already rely on threshold-based decision rules, such as determining which patients are flagged for proactive outreach, additional follow-up, or care coordination \cite{foryciarz2022evaluating}. In such cases, threshold adjustment provides a feasible mechanism for reducing disparities in decision outcomes without altering the underlying risk model. 
More broadly, ensuring model fairness and trustworthiness requires attention across the entire model development and deployment pipeline, from data curation and feature selection to model training, evaluation, and real-world use, as biases can emerge at multiple stages. 
From a practical perspective, reducing algorithmic bias may help promote more equitable decision support for patients receiving MOUD. By reducing disparities in model predictions across demographic groups, fairness-aware models may help ensure that patients with similar clinical risk profiles are identified more consistently regardless of race, ethnicity, age, or sex. This could support more equitable treatment prioritization, referral decisions, and timely interventions for patients at increased risk of premature treatment discontinuation or poor treatment retention. For example, clinicians may be better able to identify individuals who could benefit from additional counseling, case management, or closer follow-up while reducing the likelihood that certain demographic groups are systematically under- or over-identified for these supportive services. Such improvements may contribute to more equitable allocation of healthcare resources and promote more consistent access to evidence-based interventions across diverse patient populations. Nevertheless, algorithmic fairness alone cannot eliminate existing healthcare disparities. Achieving equitable care also requires addressing broader structural, socioeconomic, and healthcare system factors that influence treatment access, engagement, and outcomes. Therefore, fairness-aware predictive models should be viewed as decision support tools that complement, rather than replace, clinical judgment and broader efforts to advance health equity. 

Our assessment of model fairness and performance inconsistencies has several important limitations. 
First, from a predictive performance perspective, the overall performance achieved by the models was modest compared with that reported in some prior MOUD-related prediction studies \cite{stafford2022identifying,warren2022using}. This is largely caused by challenges inherent to outpatient treatment settings and to the limitations of routinely collected administrative data. Unlike inpatient, residential, or mixed treatment settings, outpatient MOUD care is less structured. In this context, treatment retention is influenced by dynamic and often unobserved behavioral and social factors, and sustained engagement relies heavily on patient self-management over extended periods \cite{baird2023determinants}. 
Although the TEDS-D dataset is large and nationally representative, it lacks detailed and longitudinal clinical and contextual information, such as provider behavior, social support, and individual treatment adherence, all of which are important for understanding treatment retention and outcomes. 
In the TEDS-D dataset, MOUD is recorded as a single indicator of planned medication-assisted opioid therapy and does not distinguish among methadone, buprenorphine, and naltrexone. 
Our outcome definition may also be constrained by the limited information available in the dataset and may not fully distinguish the circumstances surrounding treatment termination. 
In addition, restricting the cohort to outpatient settings increases population homogeneity and removes setting-driven variation that may otherwise inflate predictive performance in mixed-setting analyses. As a result, achieving high overall performance for more complex, behaviorally driven outcomes such as treatment retention is inherently more challenging in outpatient MOUD settings. 
Future MOUD research would benefit substantially from the availability of richer datasets that incorporate longitudinal clinical, behavioral, and social determinants of health, which may enable more accurate prediction while also supporting more nuanced fairness assessments. 
Second, although our outcome definitions—such as retention over 180 days and premature discontinuation—are consistent with prior research, they may not fully reflect the nuanced and dynamic nature of treatment engagement and outcomes. Treatment trajectories are often influenced by both personal and structural factors that are not easily captured by binary outcomes. As such, some complexity of real-world treatment behavior may be lost in this formulation, limiting the models’ ability to represent patient trajectories accurately. 
Third, the retrospective nature of our analysis, which relies on administrative discharge data, may limit the generalizability of our findings to other clinical settings or populations. Moreover, the dataset includes only facilities that report to TEDS-D, which may not reflect the full diversity of MOUD treatment settings, such as private or community-based clinics, limiting generalizability. Finally, while we evaluated bias mitigation strategies across multiple stages of the ML pipeline, our analysis does not assess how these interventions would perform when deployed in real-world clinical environments. Factors such as clinical workflow integration, provider response to model outputs, and the downstream consequences of prediction errors remain unexamined. 

Looking ahead, future research can take several important directions to further advance the development of fair ML models in the context of MOUD treatment outcome prediction. 
First, there is a continued need to develop strategies that can more effectively balance algorithmic fairness and overall predictive performance. While many current approaches seek to improve one at the expense of the other, future work could explore adaptive or context-aware algorithms that optimize both dimensions simultaneously, minimizing unintended trade-offs. 
More robust and domain-specific mitigation strategies should be explored to ensure generalizability across diverse clinical settings and populations. 
Second, evaluating the real-world impact of these mitigation strategies is critical. There is a need for prospective studies, implementation trials, and collaborations with health systems to assess how these strategies influence clinical workflows, decision-making, and patient outcomes in practice over time. Additionally, longitudinal studies could examine the durability of mitigation techniques, their acceptability among clinicians, and their integration into routine decision-making workflows. 
In addition, engaging stakeholders—including clinicians, patients, researchers, and implementation teams—through participatory design frameworks can help ensure that algorithmic tools are aligned with clinical priorities, ethically grounded, and contextually appropriate for the communities they are intended to serve. 
Finally, identifying and understanding the root causes of observed model performance gaps across patient subgroups remains a crucial area for future work. For instance, using causal inference techniques could help distinguish between variables that genuinely influence treatment outcomes and those that merely act as proxies for structural patterns in the data. By uncovering these causal relationships, researchers can design models that focus on clinically meaningful signals while minimizing the influence of spurious or confounding patterns that may distort predictions. 


\section{Conclusion}
This study systematically examined the use of ML models to predict MOUD treatment outcomes, including premature treatment discontinuation and treatment retention exceeding 180 days, with a specific focus on outpatient treatment settings. Using a large, national administrative dataset, we evaluated both predictive performance and algorithmic fairness across patient sociodemographic subgroups, and assessed the effectiveness of commonly used bias mitigation strategies spanning different stages of the ML development pipeline. 
Our analysis revealed notable discrepancies in model performance across patient subgroups defined by sensitive attributes such as race and age, as well as more complex cross-tabulated subgroups. We found that ML models optimized for overall performance can exhibit systematically different error rates and predictive patterns across sociodemographic groups, raising important concerns about fairness and equity when such models are used to inform care decisions. Our interpretation of model outputs suggests that such performance variation across subgroups is likely caused by the model learning spurious patterns in the training data. 
To address these concerns, we assessed a set of mitigation techniques aimed at improving model fairness across patient subgroups. While these strategies were able to reduce performance gaps in many cases, they often introduced trade-offs, such as shifts in prediction error between subgroups, highlighting the need for cautious implementation and continuous monitoring. 
Our evaluation of bias mitigation strategies further demonstrates that no single approach is universally optimal. Pre-processing, in-processing, and post-processing methods each offer distinct advantages and trade-offs depending on implementation constraints, technical capacity, and intended use. These results provide practical guidance for researchers and practitioners seeking to integrate fairness-aware ML into MOUD treatment planning. 
Ultimately, our findings emphasize the importance of developing ML models that are not only accurate overall but also perform well across different patient populations. Achieving this balance requires greater attention to the design, training, and evaluation of models, as well as clear guidelines for how subgroup-level performance should be monitored and reported. 
Future development efforts must integrate these principles to ensure that ML-based tools support fair and clinically meaningful decision-making for MOUD.

\backmatter











\section*{Statements and Declarations}

\subsection*{Competing Interests}
The authors have no competing interests to declare that are relevant to the content of this article and there are no financial interests.

\subsection*{Funding Acknowledgements}
This research was supported, in part, by the National Institutes of Health (NIH) under Agreement No. 1OT2OD032581 and by the U.S. National Science Foundation under Grant No. CMMI-2222670.

\subsection*{Ethics Approval}
Not applicable.

\subsection*{Consent to Participate}
Not applicable.

\subsection*{Consent for Publication}
Not applicable.

\subsection*{Availability of data and materials}
Data used in this study are openly available at \href{https://www.samhsa.gov/data/data-we-collect/teds-treatment-episode-data-set}{SAMHSA Treatment Episode Data Set (TEDS)}.


\newcounter{maintextfigure}
\newcounter{maintexttable}
\setcounter{maintextfigure}{\value{figure}}
\setcounter{maintexttable}{\value{table}}

\begin{appendices}

\setcounter{figure}{\value{maintextfigure}}
\setcounter{table}{\value{maintexttable}}
\renewcommand{\thefigure}{\arabic{figure}}
\renewcommand{\thetable}{\arabic{table}}

\section{Supplementary Results}\label{secA1}

\begin{figure}[tbp]
\centering
\includegraphics[width=0.6\textwidth]{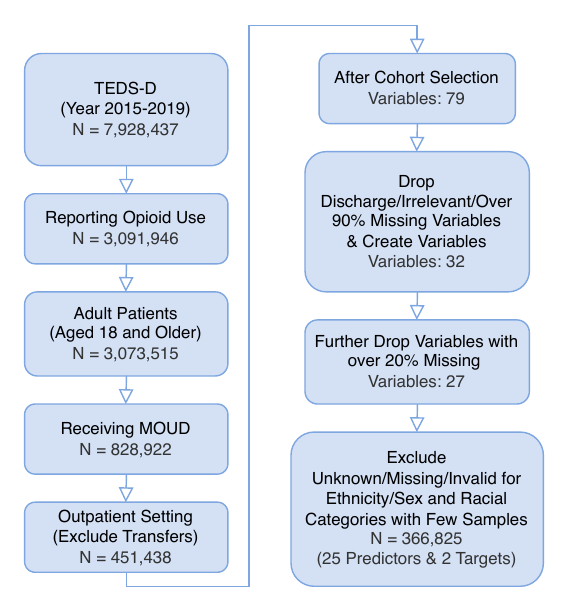}\label{fig:TEDS_data_preprocess} 
\caption{Cohort and variable selection.} 
\label{fig:cohort_variable}
\end{figure}

\subsection{Data processing}
Figure~\ref{fig:cohort_variable} illustrates the process used for cohort construction and variable selection in this study. Following this process, a total of 27 variables were included. Of these, two served as outcome variables—treatment retention longer than 180 days and premature discontinuation—leaving 25 variables as predictors for model development.

\subsection{Hyperparameter Tuning} \label{hyperparameter_tuning}
For LR, RF, and GBDT, hyperparameters were optimized using grid search with 5-fold stratified cross-validation on the training set. All candidate hyperparameter combinations in the predefined search grid were evaluated using the mean AUROC across the five validation folds. The configuration achieving the highest mean cross-validated AUROC was selected, and the corresponding model was refitted on the complete training set before evaluation on the held-out test set. 
Hyperparameter tuning for the MLP model was performed using grid search with Keras Tuner. Candidate hyperparameter configurations were evaluated on the validation set using AUROC as the optimization criterion. The hyperparameter configuration achieving the highest validation AUROC was selected for the final model. 
Results are reported with averages over 10 independent runs using different random seeds to account for variability. Best hyperparameter configurations for each model and task are summarized in Table~\ref{tab:hyperparameter}.

\subsection{Equalized Odds Results}\label{EOD_results}
Equalized odds is a fairness criterion that requires equal TPRs and FPRs across subgroups. 
In this study, we evaluated it for all four ML models across subgroups defined by race, ethnicity, and age to provide a comprehensive assessment of fairness. The results are presented in Table~\ref{tab:equalized_odds} for both premature treatment discontinuation and treatment retention longer than 180 days. 
\vspace{12pt}

\begin{sidewaystable}[tbp]
\caption{Equalized Odds Difference (mean; standard deviation in parentheses) for predicting premature discontinuation and retention longer than 180 days. The Equalized Odds Difference is defined as the maximum of the TPR difference and the FPR difference across subgroups. Lower values indicate smaller gaps between subgroups, with 0 representing perfect equalized odds. \textbf{Bold} values indicate the best results.}
\label{tab:equalized_odds}
\begin{tabular*}{\textheight}{@{\extracolsep\fill}llcccccccc@{}}
\toprule
\multirow{2}{*}{\textbf{Group}} & \multirow{2}{*}{\textbf{Method}} 
& \multicolumn{4}{c}{\textbf{Premature Discontinuation}} 
& \multicolumn{4}{c}{\textbf{Retention $> 180$ days}} \\

\cmidrule(lr){3-6} \cmidrule(l){7-10} 
&  & LR & RF & GBDT & MLP & LR & RF & GBDT & MLP \\

\midrule
\multirow{8}{*}{Race} 
& Baseline 
& \makecell{0.2967 \\ (0.0060)} & \makecell{0.2526 \\ (0.0104)} & \makecell{0.2681 \\ (0.0095)} & \makecell{0.2828 \\ (0.0410)}    
& \makecell{0.2081 \\ (0.0073)} & \makecell{0.2255 \\ (0.0048)} & \makecell{0.2354 \\ (0.0075)} & \makecell{0.2119 \\ (0.0394)}    \\
\cmidrule(lr){3-6} \cmidrule(l){7-10} 
& Reweighing   
& \makecell{0.0289 \\ (0.0069)} & \makecell{0.0452 \\ (0.0041)} & \makecell{0.0319 \\ (0.0073)} & \textbf{\makecell{0.0434 \\ (0.0270)}}    
& \makecell{0.0147 \\ (0.0043)} & \makecell{0.0987 \\ (0.0055)} & \makecell{0.0792 \\ (0.0084)} & \textbf{\makecell{0.0831 \\ (0.0471)}}    \\
\cmidrule(lr){3-6} \cmidrule(l){7-10} 
& EGR    
& \makecell{0.0108 \\ (0.0073)} & \makecell{0.0489 \\ (0.0434)} & \makecell{0.0449 \\ (0.0304)} & -- \footnotemark[5]  
& \makecell{0.0121 \\ (0.0025)} & \textbf{\makecell{0.0082 \\ (0.0080)}} & \textbf{\makecell{0.0078 \\ (0.0053)}} & -- \footnotemark[5]    \\
\cmidrule(lr){3-6} \cmidrule(l){7-10} 
& Threshold    
& \textbf{\makecell{0.0065 \\ (0.0036)}} & \textbf{\makecell{0.0295 \\ (0.0060)}} & \textbf{\makecell{0.0135 \\ (0.0053)}} & \makecell{0.2011 \\ (0.0222)}    
& \textbf{\makecell{0.0063 \\ (0.0058)}} & \makecell{0.0134 \\ (0.0051)} & \makecell{0.0085 \\ (0.0043)} & \makecell{0.2257 \\ (0.0267)}    \\

\midrule
\multirow{8}{*}{Ethnicity} 
& Baseline     
& \makecell{0.1895 \\ (0.0078)} & \makecell{0.1406 \\ (0.0089)} & \makecell{0.1709 \\ (0.0171)} & \makecell{0.1786 \\ (0.0230)} 
& \makecell{0.1437 \\ (0.0103)} & \makecell{0.1133 \\ (0.0097)} & \makecell{0.1101 \\ (0.0157)} & \makecell{0.1226 \\ (0.0235)}   \\
\cmidrule(lr){3-6} \cmidrule(l){7-10} 
& Reweighing   
& \makecell{0.0522 \\ (0.0073)} & \makecell{0.0250 \\ (0.0071)} & \makecell{0.0333 \\ (0.0132)} & \textbf{\makecell{0.0428 \\ (0.0190)}}    
& \textbf{\makecell{0.0084 \\ (0.0032)}} & \makecell{0.0379 \\ (0.0074)} & \textbf{\makecell{0.0083 \\ 0.0052}} & \textbf{\makecell{0.0245 \\ (0.0170)}}    \\
\cmidrule(lr){3-6} \cmidrule(l){7-10} 
& EGR      
& \makecell{0.0153 \\ (0.0085)} & \makecell{0.0830 \\ (0.0059)} & \makecell{0.0453 \\ (0.0309)} & --    
& \makecell{0.0118 \\ (0.0072)} & \textbf{\makecell{0.0114 \\ (0.0077)}} & \makecell{0.0123 \\ (0.0076)} & --    \\
\cmidrule(lr){3-6} \cmidrule(l){7-10} 
& Threshold    
& \textbf{\makecell{0.0099 \\ (0.0085)}} & \textbf{\makecell{0.0203 \\ (0.0082)}} & \textbf{\makecell{0.0088 \\ (0.0091)}} & \makecell{0.0975 \\ (0.0182)}    
& \makecell{0.0090 \\ (0.0073)} & \makecell{0.0182 \\ (0.0106)} & \makecell{0.0112 \\ (0.0070)} & \makecell{0.1091 \\ (0.0236)}    \\

\midrule
\multirow{8}{*}{Age} 
& Baseline    
& \makecell{0.1441 \\ (0.0100)} & \makecell{0.1726 \\ (0.0127)} & \makecell{0.1445 \\ (0.0096)} & \makecell{0.1139 \\ (0.0315)} 
& \makecell{0.5619 \\ (0.0095)} & \makecell{0.4278 \\ (0.0151)} & \makecell{0.4497 \\ (0.0152)} & \makecell{0.4842 \\ (0.0428)}    \\
\cmidrule(lr){3-6} \cmidrule(l){7-10} 
& Reweighing   
& \makecell{0.0345 \\ (0.0105)} & \makecell{0.0930 \\ (0.0096)} & \makecell{0.0457 \\ (0.0089)} & \textbf{\makecell{0.0482 \\ (0.0285)}}    
& \makecell{0.0217 \\ (0.0043)} & \makecell{0.0497 \\ (0.0057)} & \makecell{0.0447 \\ (0.0103)} & \textbf{\makecell{0.0607 \\ (0.0297)}}    \\
\cmidrule(lr){3-6} \cmidrule(l){7-10} 
& EGR          
& \makecell{0.0183 \\ (0.0078)} & \makecell{0.1035 \\ (0.0126)} & \makecell{0.0459 \\ (0.0351)} & --    
& \makecell{0.0212 \\ (0.0093)} & \textbf{\makecell{0.0294 \\ (0.0033)}} & \makecell{0.0273 \\ (0.0059)} & --   \\
\cmidrule(lr){3-6} \cmidrule(l){7-10} 
& Threshold  
& \textbf{\makecell{0.0110 \\ (0.0037)}} & \textbf{\makecell{0.0137 \\ (0.0064)}} & \textbf{\makecell{0.0116 \\ (0.0053)}} & \makecell{0.0579 \\ (0.0123)}    
& \textbf{\makecell{0.0109 \\ (0.0050)}} & \makecell{0.0401 \\ (0.0061)} & \textbf{\makecell{0.0170 \\ (0.0062)}} & \makecell{0.4529 \\ (0.0344)}    \\

\bottomrule
\end{tabular*}
\footnotetext[5]{Results for MLP with EGR are not reported, as EGR requires models that support stable, per-sample reweighting during training. MLP are non-convex and typically do not support per-sample reweighting in a stable and standardized manner as required.} 
\end{sidewaystable}

\begin{sidewaystable}[tbp]
\caption{Best hyperparameter configurations.}\label{tab:hyperparameter}%
\begin{tabular*}{\textwidth}{@{\extracolsep\fill}lllllllll@{}}
\toprule
& \multicolumn{2}{c}{\textbf{LR}} & \multicolumn{2}{c}{\textbf{RF}} & \multicolumn{2}{c}{\textbf{GBDT}} & \multicolumn{2}{c}{\textbf{MLP}} \\

\cmidrule(lr){2-3} \cmidrule(lr){4-5} \cmidrule(lr){6-7} \cmidrule(lr){8-9}
\multirow{5}{*}{Premature Discontinuation} 
  &  solver        &  saga     & n\_estimators      & 500  & l2\_regularization  & 0.5  &  hidden layer  & 32 (ReLU)  \\
  &  penalty       &  l1       & max\_depth         & 20   & learning\_rate      & 0.01 &  output layer  & 1 (Sigmoid) \\
  &  C             &  100      & min\_samples\_split & 20   & max\_iter           & 500  &  epoch (batch size) & 10 (64) \\
  &  max\_iter     &  50       &                   &      & max\_leaf\_nodes     & 200  &  optimizer & Adam \\
  &  class\_weight &  balanced &                   &      &  &  & learning rate & 0.001  \\
  
\midrule
\multirow{5}{*}{Retention $>$ 180 days} 
  &  solver        &  saga     & n\_estimators      & 500  & l2\_regularization  & 2.0  &  hidden layer  & 32 (ReLU)  \\
  &  penalty       &  l2       & max\_depth         & 20   & learning\_rate      & 0.01 &  output layer  & 1 (Sigmoid) \\
  &  C             &  10      & min\_samples\_split & 20   & max\_iter           & 500  &  epoch (batch size) & 10 (64) \\
  &  max\_iter     &  50       &                   &      & max\_leaf\_nodes     & 200  &  optimizer & Adam \\
  &  class\_weight &  None &                   &      &  &  & learning rate & 0.001  \\
\bottomrule
\end{tabular*}
\end{sidewaystable}

\subsection{Accuracy and Fairness Trade-off}
Table~\ref{tab:overall_performance_dropout_los} reports overall model accuracy for both tasks. When compared with the fairness metrics presented in Tables~\ref{tab:group_fairness_dropout} and~\ref{tab:group_fairness_los}, the results reveal a clear trade-off between accuracy and fairness. 

\begin{sidewaystable}[tbp]
\caption{Model accuracy (mean; standard deviation in parentheses) for predicting premature treatment discontinuation and treatment retention. \textbf{Bold} values indicate the best results.}
\label{tab:overall_performance_dropout_los}
\begin{tabular*}{\textwidth}{@{\extracolsep\fill}llcccccccc@{}}
\toprule
\multirow{2}{*}{\textbf{Group}} & \multirow{2}{*}{\textbf{Method}} & \multicolumn{4}{c}{\textbf{Premature Discontinuation}} & \multicolumn{4}{c}{\textbf{Retention $>$ 180 days}} \\

\cmidrule(lr){3-6} \cmidrule(lr){7-10}
&  & LR & RF & GBDT & MLP & LR & RF & GBDT & MLP  \\
\midrule
\multirow{8}{*}{Race} 
& Baseline 
& \makecell{0.6104 \\ (0.0011)} & \makecell{0.6230 \\ (0.0014)} & \makecell{0.6263 \\ (0.0014)} & \makecell{0.6065 \\ (0.0061)}    
& \makecell{0.6434 \\ (0.0010)} & \makecell{0.6551 \\ (0.0007)} & \makecell{0.6587 \\ (0.0009)} & \makecell{0.6469 \\ (0.0014)}        \\
\cmidrule(lr){3-6} \cmidrule(lr){7-10}
& Reweighing   
& \makecell{0.6058 \\ (0.0011)} & \makecell{0.6218 \\ (0.0013)} & \makecell{0.6251 \\ (0.0014)} & \makecell{0.6193 \\ (0.0063)}    
& \makecell{0.6416 \\ (0.0007)} & \makecell{0.6544 \\ (0.0007)} & \makecell{0.6576 \\ (0.0007)} & \makecell{0.6521 \\ (0.0018)}        \\
\cmidrule(lr){3-6} \cmidrule(lr){7-10}
& EGR    
& \makecell{0.6099 \\ (0.0022)} & \makecell{0.6316 \\ (0.0021)} & \makecell{0.6340 \\ (0.0010)} & -- \footnotemark[5]  
& \makecell{0.6416 \\ (0.0008)} & \makecell{0.6538 \\ (0.0009)} & \makecell{0.6568 \\ (0.0007)} & -- \footnotemark[5]                     \\
\cmidrule(lr){3-6} \cmidrule(lr){7-10}
& Threshold    
& \makecell{0.6031 \\ (0.0016)} & \makecell{0.6243 \\ (0.0014)} & \makecell{0.6241 \\ (0.0026)} & \makecell{0.6155 \\ (0.0036)}    
& \makecell{0.6086 \\ (0.0055)} & \makecell{0.6175 \\ (0.0029)} & \makecell{0.6264 \\ (0.0027)} & \makecell{0.6188 \\ (0.0055)}        \\

\midrule
\multirow{8}{*}{Ethnicity} 
& Baseline     
& \makecell{0.6104 \\ (0.0011)} & \makecell{0.6230 \\ (0.0014)} & \makecell{0.6263 \\ (0.0014)} & \makecell{0.6065 \\ (0.0061)}    
& \makecell{0.6434 \\ (0.0010)} & \makecell{0.6551 \\ (0.0007)} & \makecell{0.6587 \\ (0.0009)} & \makecell{0.6469 \\ (0.0014)}        \\
\cmidrule(lr){3-6} \cmidrule(lr){7-10}
& Reweighing   
& \makecell{0.6095 \\ (0.0013)} & \makecell{0.6225 \\ (0.0013)} & \makecell{0.6255 \\ (0.0012)} & \makecell{0.6209 \\ (0.0050)}    
& \makecell{0.6429 \\ (0.0010)} & \makecell{0.6549 \\ (0.0010)} & \makecell{0.6584 \\ (0.0009)} & \makecell{0.6527 \\ (0.0012)}        \\
\cmidrule(lr){3-6} \cmidrule(lr){7-10}
& EGR      
& \makecell{0.6104 \\ (0.0012)} & \makecell{0.6332 \\ (0.0014)} & \makecell{0.6347 \\ (0.0012)} & --   
& \makecell{0.6434 \\ (0.0010)} & \makecell{0.6550 \\ (0.0008)} & \makecell{0.6586 \\ (0.0008)} & --   \\
\cmidrule(lr){3-6} \cmidrule(lr){7-10}
& Threshold    
& \makecell{0.6062 \\ (0.0015)} & \makecell{0.6259 \\ (0.0014)} & \makecell{0.6276 \\ (0.0013)} & \makecell{0.6230 \\ (0.0017)}     
& \makecell{0.5957 \\ (0.0049)} & \makecell{0.6185 \\ (0.0024)} & \makecell{0.6283 \\ (0.0040)} & \makecell{0.6148 \\ (0.0103)}       \\

\midrule
\multirow{8}{*}{Age} 
& Baseline    
& \makecell{0.6104 \\ (0.0011)} & \makecell{0.6230 \\ (0.0014)} & \makecell{0.6263 \\ (0.0014)} & \makecell{0.6065 \\ (0.0061)}    
& \makecell{0.6434 \\ (0.0010)} & \makecell{0.6551 \\ (0.0007)} & \makecell{0.6587 \\ (0.0009)} & \makecell{0.6469 \\ (0.0014)}       \\
\cmidrule(lr){3-6} \cmidrule(lr){7-10}
& Reweighing   
& \makecell{0.6097 \\ (0.0010)} & \makecell{0.6233 \\ (0.0011)} & \makecell{0.6264 \\ (0.0012)} & \makecell{0.6222 \\ (0.0047)}    
& \makecell{0.6378 \\ (0.0007)} & \makecell{0.6515 \\ (0.0006)} & \makecell{0.6539 \\ (0.0008)} & \makecell{0.6481 \\ (0.0015)}       \\
\cmidrule(lr){3-6} \cmidrule(lr){7-10}
& EGR          
& \makecell{0.6105 \\ (0.0011)} & \makecell{0.6333 \\ (0.0014)} & \makecell{0.6348 \\ (0.0007)} & --   
& \makecell{0.6241 \\ (0.0102)} & \makecell{0.6520 \\ (0.0009)} & \makecell{0.6542 \\ (0.0008)} & --  \\
\cmidrule(lr){3-6} \cmidrule(lr){7-10}
& Threshold  
& \makecell{0.5954 \\ (0.0029)} & \makecell{0.6216 \\ (0.0025)} & \makecell{0.6257 \\ (0.0027)} & \makecell{0.6212 \\ (0.0040)}   
& \makecell{0.5705 \\ (0.0064)} & \makecell{0.6169 \\ (0.0039)} & \makecell{0.6144 \\ (0.0074)} & \makecell{0.6017 \\ (0.0078)}       \\

\bottomrule
\end{tabular*}
\footnotetext[5]{Results for MLP with EGR are not reported, as EGR requires models that support stable, per-sample reweighting during training. MLP are non-convex and typically do not support per-sample reweighting in a stable and standardized manner as required.} 
\end{sidewaystable}

\subsection{AUROC Results Across Subgroups \label{subgruop_AUROC}}
AUROC results across subgroups defined by race, ethnicity, and age are presented in Table~\ref{tab:AUROC_subgroups}. Overall, AUROC values were generally comparable across demographic groups, although modest differences were observed. However, disparities in TPRs and FPRs remained at the selected classification threshold. Accordingly, the Threshold Optimizer was employed to mitigate these threshold-dependent disparities.
\vspace{12pt}

\begin{sidewaystable}[htb]
\centering
\small
\begin{tabular*}{0.82\textwidth}{@{}llcccc@{}}
\toprule
& & LR & RF & GBDT & MLP \\

\midrule
\multirow{10}{*}{Premature Discontinuation} 
& Race & & \\
& \quad White & 0.6566 (0.0011) & 0.6865 (0.0013) & 0.6895 (0.0013) & 0.6651 (0.0024) \\
& \quad Black & 0.6291 (0.0033) & 0.6706 (0.0022) & 0.6729 (0.0024) & 0.6412 (0.0043) \\
& Ethnicity & & \\
& \quad Non-Hispanic & 0.6535 (0.0011) & 0.6840 (0.0013) & 0.6872 (0.0013) & 0.6626 (0.0027) \\
& \quad Hispanic & 0.6793 (0.0068) & 0.7072 (0.0049) & 0.7062 (0.0055) & 0.6828 (0.0056) \\
& Age & & \\
& \quad 18-24 & 0.6455 (0.0038) & 0.6780 (0.0047) & 0.6805 (0.0049) & 0.6543 (0.0049) \\
& \quad 25-54 & 0.6540 (0.0013) & 0.6843 (0.0011) & 0.6869 (0.0012) & 0.6629 (0.0025) \\
& \quad $>$ 55 & 0.6771 (0.0037) & 0.7062 (0.0042) & 0.7106 (0.0034) & 0.6829 (0.0041) \\

\midrule
\multirow{10}{*}{Retention $>$ 180 days} 
& Race & & \\
& \quad White & 0.6531 (0.0010) & 0.6759 (0.0012) & 0.6796 (0.0011) & 0.6626 (0.0014) \\
& \quad Black & 0.6600 (0.0039) & 0.6900 (0.0040) & 0.6943 (0.0044) & 0.6719 (0.0037) \\
& Ethnicity & & \\
& \quad Non-Hispanic & 0.6562 (0.0013) & 0.6793 (0.0015) & 0.6833 (0.0013) & 0.6659 (0.0015) \\
& \quad Hispanic & 0.6467 (0.0063) & 0.6788 (0.0054) & 0.6814 (0.0060) & 0.6576 (0.0065) \\
& Age & & \\
& \quad 18-24 & 0.6465 (0.0018) & 0.6682 (0.0022) & 0.6743 (0.0022) & 0.6542 (0.0034) \\
& \quad 25-54 & 0.6519 (0.0011) & 0.6768 (0.0013) & 0.6804 (0.0012) & 0.6621 (0.0012) \\
& \quad $>$ 55 & 0.6241 (0.0027) & 0.6578 (0.0033) & 0.6610 (0.0034) & 0.6416 (0.0041) \\

\bottomrule
\end{tabular*}
\caption{AUROC results for predicting premature discontinuation and treatment retention $>$ 180 days.}
\label{tab:AUROC_subgroups}
\end{sidewaystable}

\subsection{Ablation Results}
All four ML models were also trained without including sensitive attributes as input features. As shown in Table \ref{tab:ablation} for predicting premature discontinuation and Table \ref{tab:ablation2} for treatment retention over 180 days, overall prediction performance remained largely unchanged. However, performance gaps across patient subgroups persisted, indicating that excluding these attributes does not eliminate subgroup-level differences and that models may have relied on correlated features as proxies for prediction.

\subsection{Additional Results}
\begin{figure*}[tbp]
\centering
\includegraphics[width=0.98\textwidth]{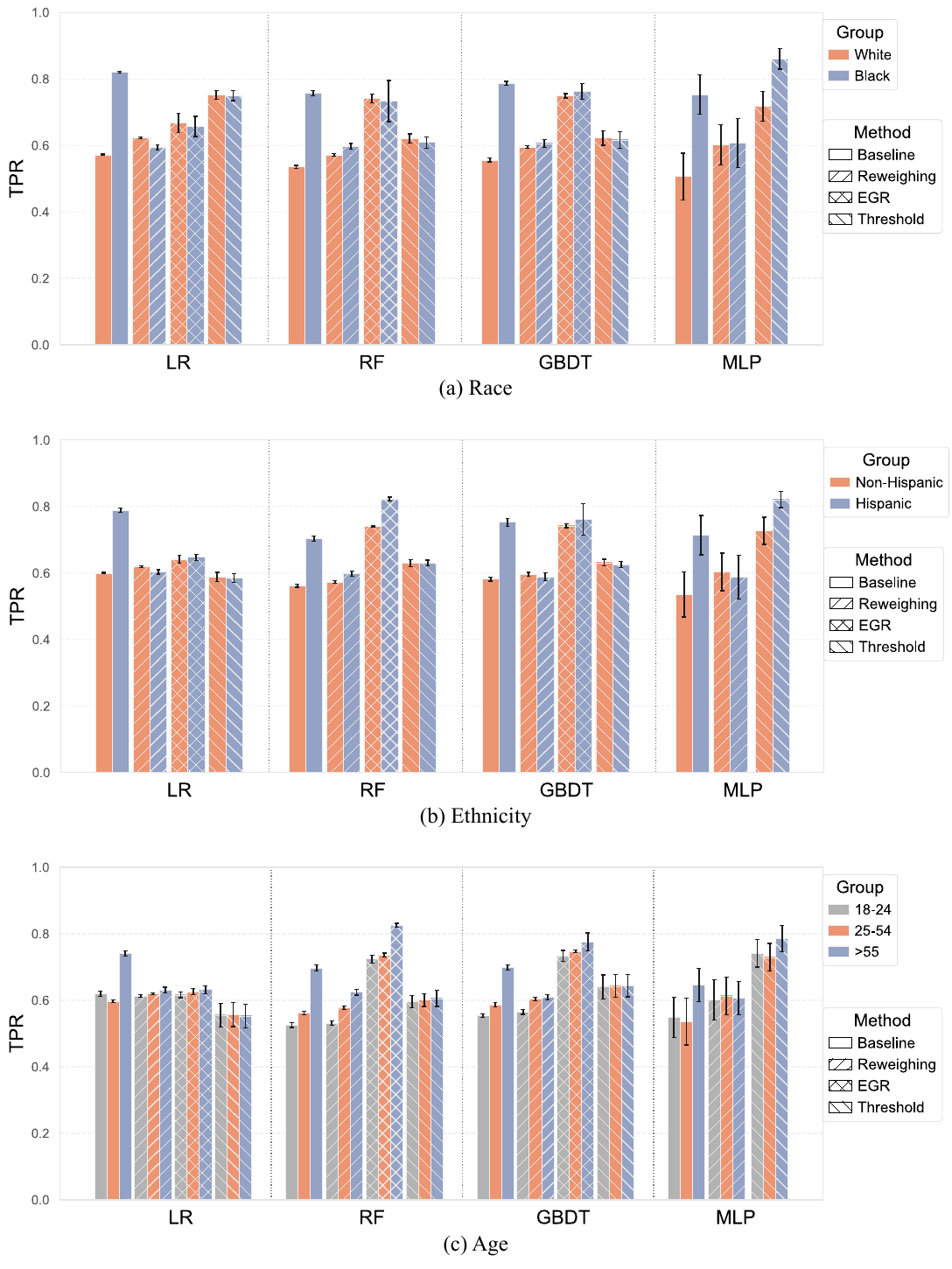}
\caption{TPRs across different subgroups are shown for three bias mitigation methods—Reweighing (pre-processing), EGR (in-processing), and Threshold (post-processing)—in predicting premature treatment discontinuation, compared to a baseline without fairness considerations. Results are stratified by (a) race, (b) ethnicity, and (c) age. Bars represent the mean TPR values over 10 independent runs, with error bars indicating the standard deviation across these runs.} 
\label{fig:dropout_bias_mitigation_TPR}
\end{figure*}

\begin{figure*}[tbp]
\centering
\includegraphics[width=0.98\textwidth]{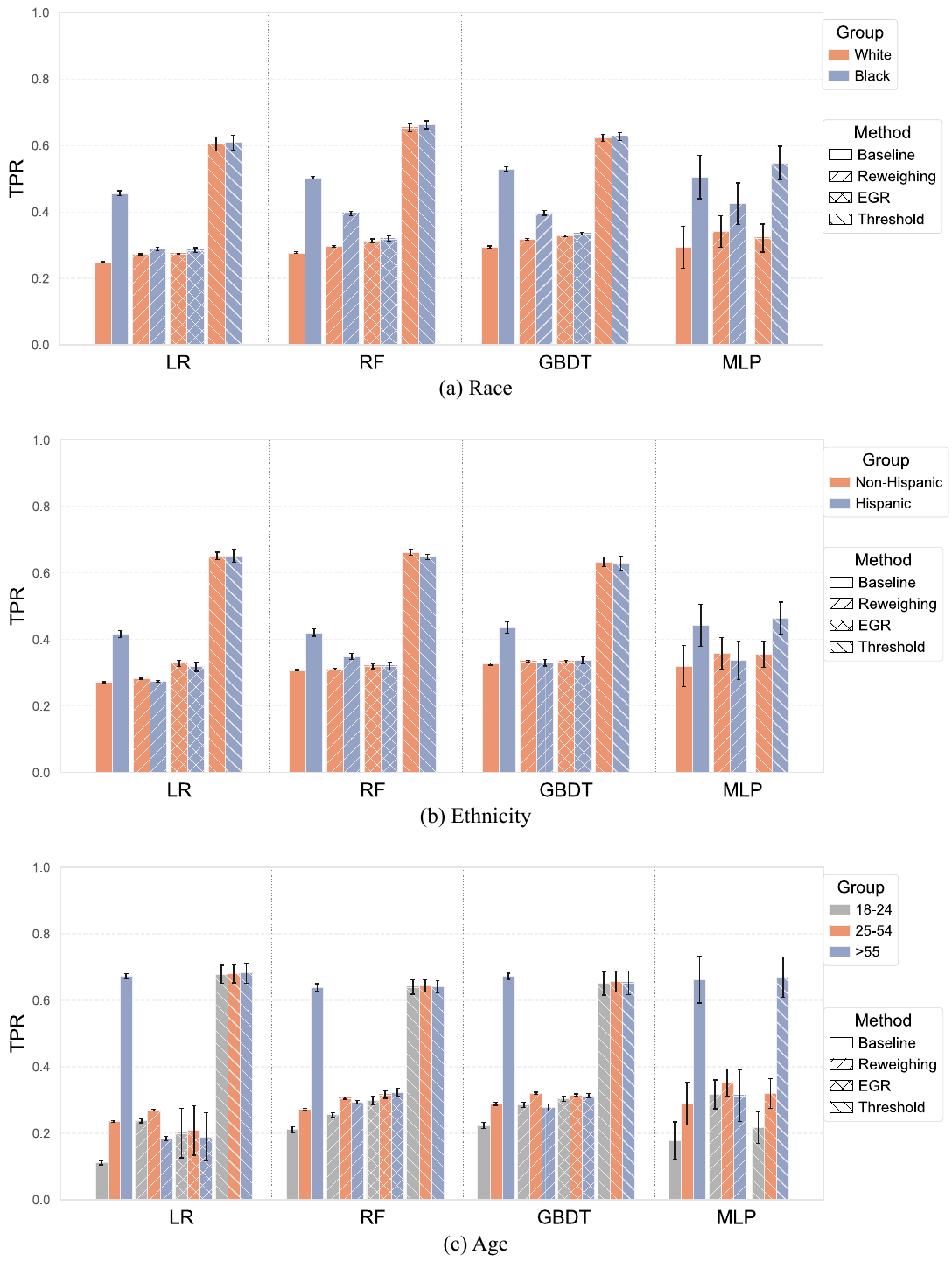}
\caption{TPRs across different subgroups are shown for three bias mitigation methods—Reweighing (pre-processing), EGR (in-processing), and Threshold (post-processing)—in predicting treatment retention $>$ 180 days, compared to a baseline without fairness considerations. Results are stratified by (a) race, (b) ethnicity, and (c) age. Bars represent the mean TPR values over 10 independent runs, with error bars indicating the standard deviation across these runs.} 
\label{fig:los_bias_mitigation_TPR}
\end{figure*}

Figure \ref{fig:dropout_bias_mitigation_TPR} presents the TPR for each bias mitigation strategy compared to the baseline when predicting premature discontinuation. 
All three methods substantially reduced TPR gaps across patient subgroups. A similar pattern is observed in Figure \ref{fig:los_bias_mitigation_TPR} for predictions of treatment retention beyond 180 days. 
Tables \ref{tab:all_FPR_dropout} and \ref{tab:all_TPR_dropout} present detailed FPR and TPR results for each ML model across all patient subgroups for predicting premature discontinuation. Similarly, Tables \ref{tab:all_FPR_los} and \ref{tab:all_TPR_los} report all FPR and TPR results for predicting treatment retention longer than 180 days.

\begin{sidewaystable}[tbp]
\caption{Comparison of model performance with and without sensitive attribute as a feature for predicting premature treatment discontinuation.}
\label{tab:ablation}
\small
\begin{tabular*}{\textheight}{@{\extracolsep\fill}llcccc@{\extracolsep\fill}}
\toprule
& & LR & RF & GBDT & MLP \\

\midrule
\multirow{3}{*}{With Race} 
& Overall (AUROC) & 0.6562 (0.0011) & 0.6863 (0.0013) & 0.6892 (0.0013) & 0.6647 (0.0025) \\
& \quad White (FPR) & 0.3545 (0.0031) & 0.2842 (0.0047) & 0.3011 (0.0064) & 0.2878 (0.0666) \\
& \quad Black (FPR) & 0.6512 (0.0067) & 0.5368 (0.0108) & 0.5692 (0.0085) & 0.5705 (0.0702) \\

\cmidrule(l){2-6} 
\multirow{3}{*}{Without Race} 
& Overall (AUROC) & 0.6548 (0.0011) & 0.6847 (0.0013) & 0.6881 (0.0013) & 0.6626 (0.0024) \\
& \quad White (FPR) & 0.3882 (0.0035) & 0.3091 (0.0040) & 0.3191 (0.0052) & 0.3142 (0.0698) \\
& \quad Black (FPR) & 0.5583 (0.0088) & 0.4522 (0.0046) & 0.4746 (0.0065) & 0.4363 (0.0857) \\

\midrule
\multirow{3}{*}{With Ethnicity} 
& Overall (AUROC) & 0.6562 (0.0011) & 0.6863 (0.0013) & 0.6892 (0.0013) & 0.6647 (0.0025) \\
& \quad Non-Hispanic (FPR) & 0.3839 (0.0033) & 0.3116 (0.0045) & 0.3277 (0.0067) & 0.3160 (0.0653)  \\
& \quad Hispanic (FPR) & 0.5565 (0.0085) & 0.4225 (0.0143) & 0.4824 (0.0168) & 0.4775 (0.0731) \\

\cmidrule(l){2-6} 
\multirow{3}{*}{Without Ethnicity} 
& Overall (AUROC) & 0.6560 (0.0012) & 0.6858 (0.0014) & 0.6890 (0.0014) & 0.6647 (0.0019) \\
& \quad Non-Hispanic (FPR) & 0.3947 (0.0030) & 0.3201 (0.0039) & 0.3367 (0.0058) & 0.3247 (0.0667) \\
& \quad Hispanic (FPR) & 0.4714 (0.0077) & 0.3555 (0.0135) & 0.3824 (0.0122) & 0.3659 (0.0700) \\

\midrule
\multirow{3}{*}{With Age} 
& Overall (AUROC) & 0.6562 (0.0011) & 0.6863 (0.0013) & 0.6892 (0.0013) & 0.6647 (0.0025) \\
& \quad 18-24 (FPR) & 0.4193 (0.0068) & 0.2850 (0.0086) & 0.3115 (0.0090) & 0.3394 (0.0602) \\
& \quad 25-54 (FPR) & 0.3808 (0.0041) & 0.3116 (0.0049) & 0.3330 (0.0059) & 0.3175 (0.0684) \\
& \quad $>$ 55 (FPR) & 0.4926 (0.0082) & 0.4157 (0.0080) & 0.4076 (0.0090) & 0.3906 (0.0563) \\

\cmidrule(l){2-6} 
\multirow{3}{*}{Without Age} 
& Overall (AUROC) & 0.6559 (0.0012) & 0.6862 (0.0012) & 0.6891 (0.0014) & 0.6651 (0.0017) \\
& \quad 18-24 (FPR) & 0.3355 (0.0055) & 0.2825 (0.0083) & 0.2958 (0.0073) & 0.2745 (0.0599) \\
& \quad 25-54 (FPR) & 0.3820 (0.0036) & 0.3105 (0.0044) & 0.3249 (0.0054) & 0.3071 (0.0618) \\
& \quad $>$ 55 (FPR) & 0.5675 (0.0050) & 0.4643 (0.0059) & 0.4779 (0.0079) & 0.4866 (0.0664)  \\

\bottomrule
\end{tabular*}
\end{sidewaystable}

\begin{sidewaystable}[tbp]
\caption{Comparison of model performance with and without sensitive attribute as a feature for predicting treatment retention.}
\label{tab:ablation2}
\small
\begin{tabular*}{\textheight}{@{\extracolsep\fill}llcccc@{\extracolsep\fill}}
\toprule
& & LR & RF & GBDT & MLP \\

\midrule
\multirow{3}{*}{With Race} 
& Overall (AUROC) & 0.6559 (0.0011) & 0.6797 (0.0011) & 0.6836 (0.0010) & 0.6657 (0.0014) \\
& \quad White (FPR) & 0.1107 (0.0021) & 0.1114 (0.0016) & 0.1155 (0.0020) & 0.1288 (0.0397) \\
& \quad Black (FPR) & 0.2459 (0.0050) & 0.2495 (0.0053) & 0.2656 (0.0068) & 0.2701 (0.0465) \\

\cmidrule(l){2-6}
\multirow{3}{*}{Without Race} 
& Overall (AUROC) & 0.6556 (0.0011) & 0.6784 (0.0012) & 0.6823 (0.0009) & 0.6650 (0.0015) \\
& \quad White (FPR) & 0.1179 (0.0021) & 0.1182 (0.0018) & 0.1216 (0.0017) & 0.1391 (0.0437) \\
& \quad Black (FPR) & 0.1989 (0.0032) & 0.2236 (0.0051) & 0.2336 (0.0060) & 0.2259 (0.0522) \\

\midrule
\multirow{3}{*}{With Ethnicity} 
& Overall (AUROC) & 0.6559 (0.0011) & 0.6797 (0.0011) & 0.6836 (0.0010) & 0.6657 (0.0014) \\
& \quad Non-Hispanic (FPR) & 0.1233 (0.0021) & 0.1264 (0.0016) & 0.1319 (0.0018) & 0.1471 (0.0401) \\
& \quad Hispanic (FPR) & 0.2264 (0.0092) & 0.2023 (0.0102) & 0.2124 (0.0139) & 0.2422 (0.0505) \\

\cmidrule(l){2-6}
\multirow{3}{*}{Without Ethnicity} 
& Overall (AUROC) & 0.6559 (0.0011) & 0.6795 (0.0012) & 0.6835 (0.0010) & 0.6651 (0.0012) \\
& \quad Non-Hispanic (FPR) & 0.1269 (0.0021) & 0.1281 (0.0018) & 0.1332 (0.0019) & 0.1475 (0.0372) \\
& \quad Hispanic (FPR) & 0.1765 (0.0041) & 0.1944 (0.0067) & 0.1947 (0.0048) & 0.2051 (0.0499) \\

\midrule
\multirow{3}{*}{With Age} 
& Overall (AUROC) & 0.6559 (0.0011) & 0.6797 (0.0011) & 0.6836 (0.0010) & 0.6657 (0.0014) \\
& \quad 18-24 (FPR) & 0.1042 (0.0020) &  0.1084 (0.0022) &  0.1129 (0.0026) &  0.1320 (0.0403) \\
& \quad 25-54 (FPR) & 0.1042 (0.0020) &  0.1084 (0.0022) &  0.1129 (0.0026) &  0.1320 (0.0403) \\
& \quad $>$ 55 (FPR) & 0.4901 (0.0096) &  0.4131 (0.0110) &  0.4414 (0.0073) &  0.4584 (0.0693) \\

\cmidrule(l){2-6}
\multirow{3}{*}{Without Age} 
& Overall (AUROC) & 0.6481 (0.0010) & 0.6743 (0.0010) & 0.6773 (0.0009) & 0.6580 (0.0019) \\
& \quad 18-24 (FPR) & 0.1040 (0.0032) & 0.1130 (0.0028) & 0.1079 (0.0044) & 0.1254 (0.0471) \\
& \quad 25-54 (FPR) & 0.1120 (0.0021) & 0.1244 (0.0020) & 0.1193 (0.0020) & 0.1356 (0.0418) \\
& \quad $>$ >55 (FPR) & 0.2507 (0.0066) & 0.2747 (0.0042) & 0.2817 (0.0050) & 0.2769 (0.0518) \\

\bottomrule
\end{tabular*}
\end{sidewaystable}

\begin{table}[htb]
\centering
\small
\begin{tabular}{@{}llcccc@{}}
\toprule
& & LR & RF & GBDT & MLP \\

\midrule
\multirow{10}{*}{Baseline} 
& Race & & \\
& \quad White & 0.3545 (0.0031) & 0.2842 (0.0047) & 0.3011 (0.0064) & 0.2878  (0.0666)  \\
& \quad Black & 0.6512 (0.0067) & 0.5368 (0.0108) & 0.5692 (0.0085) & 0.5705 (0.0702) \\
& Ethnicity & & \\
& \quad Non-Hispanic & 0.3839 (0.0033) & 0.3116 (0.0045) & 0.3277 (0.0067) & 0.3160 (0.0653) \\
& \quad Hispanic & 0.5565 (0.0085) & 0.4225 (0.0143) & 0.4824 (0.0168) & 0.4775 (0.0731) \\
& Age & & \\
& \quad 18-24 & 0.4194 (0.0068) & 0.2850 (0.0086) & 0.3115 (0.0090)  & 0.3394 (0.0602) \\
& \quad 25-54 & 0.3808 (0.0041) & 0.3116 (0.0049) & 0.3330 (0.0059) & 0.3175 (0.0684) \\
& \quad $>$ 55 & 0.4926 (0.0082) & 0.4157 (0.0080) & 0.4076 (0.0090) & 0.3906 (0.0563)  \\

\midrule
\multirow{10}{*}{Reweighing} 
& Race & & \\
& \quad White & 0.4071 (0.0028) & 0.3176 (0.0040) & 0.3373 (0.0047) & 0.3559 (0.0592)  \\
& \quad Black & 0.4207 (0.0076) & 0.3628 (0.0058) & 0.3692 (0.0095) & 0.3896 (0.0710) \\
& Ethnicity & & \\
& \quad Non-Hispanic & 0.4042 (0.0032) & 0.3219 (0.0040) & 0.3415 (0.0059) & 0.3599 (0.0565)  \\
& \quad Hispanic & 0.3521 (0.0078) & 0.3104 (0.0101) & 0.3081 (0.0146) & 0.3171 (0.0663) \\
& Age & & \\
& \quad 18-24 & 0.4133 (0.0066) & 0.2919 (0.0083) & 0.3212 (0.0085) & 0.3644 (0.0581) \\
& \quad 25-54 & 0.4048 (0.0034) & 0.3261 (0.0050) & 0.3492 (0.0066) & 0.3701 (0.0566) \\
& \quad $>$ 55 & 0.3790 (0.0081) & 0.3352 (0.0079) & 0.3115 (0.0082) & 0.3241 (0.0497)  \\

\midrule
\multirow{10}{*}{EGR} 
& Race & & \\
& \quad White & 0.4953 (0.0349) & 0.4940 (0.0139) & 0.4990 (0.0082)  & -- \\
& \quad Black & 0.5061 (0.0349) & 0.5160 (0.0684) & 0.5424 (0.0304) & -- \\
& Ethnicity & & \\
& \quad Non-Hispanic & 0.4304 (0.0147) & 0.4956 (0.0014) & 0.4937 (0.0082)  & -- \\
& \quad Hispanic & 0.4151 (0.0211) & 0.5697 (0.0077) & 0.4950 (0.0616) & -- \\
& Age & & \\
& \quad 18-24 & 0.4171 (0.0121) & 0.4846 (0.0133) & 0.4928 (0.0151) & -- \\
& \quad 25-54 & 0.4114 (0.0092) & 0.4903 (0.0058) & 0.4993 (0.0051) & -- \\
& \quad $>$ 55 & 0.4042 (0.0119) & 0.5806 (0.0133) & 0.5078 (0.0347) & -- \\

\midrule
\multirow{10}{*}{Threshold} 
& Race & & \\
& \quad White & 0.5719 (0.0137) & 0.3652 (0.0146) & 0.3706 (0.0244) & 0.4732 (0.0524) \\
& \quad Black & 0.5745 (0.0131) & 0.3958 (0.0194) & 0.3842 (0.0215) & 0.6722 (0.0441) \\
& Ethnicity & & \\
& \quad Non-Hispanic & 0.3722 (0.0142) & 0.3775 (0.0098) & 0.3756 (0.0109) & 0.4917 (0.0491) \\
& \quad Hispanic & 0.3674 (0.0183) & 0.3981 (0.0115) & 0.3807 (0.0156) & 0.5856 (0.0360) \\
& Age & & \\
& \quad 18-24 & 0.3583 (0.0381) & 0.3518 (0.0206) & 0.3956 (0.0327) & 0.5166 (0.0503) \\
& \quad 25-54 & 0.3585 (0.0369) & 0.3536 (0.0181) & 0.3936 (0.0354) & 0.4959 (0.0515) \\
& \quad $>$ 55 & 0.3633 (0.0388) & 0.3616 (0.0243) & 0.3962 (0.0329) & 0.5275 (0.0561) \\

\bottomrule
\end{tabular}
\caption{All FPR results for predicting premature treatment discontinuation.}
\label{tab:all_FPR_dropout}
\end{table}

\begin{table}[htb]
\centering
\small
\begin{tabular}{@{}llcccc@{}}
\toprule
& & LR & RF & GBDT & MLP \\

\midrule
\multirow{10}{*}{Baseline} 
& Race & & \\
& \quad White & 0.5718 (0.0025) &  0.5352 (0.0047) &  0.5559 (0.0057) &  0.5069 (0.0700)  \\
& \quad Black & 0.8209 (0.0027) &  0.7564 (0.0078) &  0.7863 (0.0063) &   0.7532 (0.0588) \\
& Ethnicity & & \\
& \quad Non-Hispanic & 0.5998 (0.0025) & 0.5623 (0.0048) & 0.5821 (0.0064) & 0.5352 (0.0673) \\
& \quad Hispanic & 0.7893 (0.0064) & 0.7029 (0.0080) & 0.7530 (0.0118) & 0.7139 (0.0594) \\
& Age & & \\
& \quad 18-24 & 0.6189 (0.0079) & 0.5249 (0.0069) & 0.5537 (0.0060) & 0.5483 (0.0605) \\
& \quad 25-54 & 0.5970 (0.0033) & 0.5623 (0.0053) & 0.5869 (0.0058) & 0.5365 (0.0707) \\
& \quad $>$ 55 & 0.7411 (0.0080) & 0.6976 (0.0090) & 0.6982 (0.0079) & 0.6454  (0.0499) \\

\midrule
\multirow{10}{*}{Reweighing} 
& Race & & \\
& \quad White & 0.6233 (0.0024) &  0.5712 (0.0037) &  0.5951 (0.0035) &  0.6013 (0.0601)  \\
& \quad Black & 0.5945 (0.0076) &  0.5972 (0.0085) &  0.6057 (0.0115) &  0.6068 (0.0744) \\
& Ethnicity & & \\
& \quad Non-Hispanic & 0.6195 (0.0024) & 0.5727 (0.0044) & 0.5963 (0.0057) & 0.6034 (0.0566)  \\
& \quad Hispanic & 0.6034 (0.0068) & 0.5977 (0.0081) & 0.5887 (0.0123) & 0.5876 (0.0663) \\
& Age & & \\
& \quad 18-24 & 0.6125 (0.0039) & 0.5312 (0.0064) & 0.5649 (0.0073) & 0.6012 (0.0597) \\
& \quad 25-54 & 0.6202 (0.0023) & 0.5775 (0.0050) & 0.6036 (0.0059) & 0.6133 (0.0561) \\
& \quad $>$ 55 & 0.6312 (0.0081) & 0.6242 (0.0088) & 0.6088 (0.0078) & 0.6072  (0.0498) \\

\midrule
\multirow{10}{*}{EGR} 
& Race & & \\
& \quad White & 0.6673 (0.0292) &  0.7413 (0.0133) &  0.7491 (0.0068) &  -- \\
& \quad Black & 0.6572 (0.0303) &  0.7333 (0.0620) &  0.7626 (0.0239) &  -- \\
& Ethnicity & & \\
& \quad Non-Hispanic & 0.6410 (0.0117) & 0.7396 (0.0022) & 0.7420 (0.0059)  & -- \\
& \quad Hispanic & 0.6477 (0.0092) & 0.8227 (0.0058) & 0.7613 (0.0468) & -- \\
& Age & & \\
& \quad 18-24 & 0.6161 (0.0081) & 0.7241 (0.0118) & 0.7336 (0.0166) & -- \\
& \quad 25-54 & 0.6261 (0.0088) & 0.7361 (0.0061) & 0.7465 (0.0039) & -- \\
& \quad $>$ 55 & 0.6322 (0.0118) & 0.8256 (0.0059) & 0.7753 (0.0271) & -- \\

\midrule
\multirow{10}{*}{Threshold} 
& Race & & \\
& \quad White & 0.7510 (0.0124) &  0.6209 (0.0137) &  0.6224 (0.0217) &  0.7141 (0.0465) \\
& \quad Black & 0.7501 (0.0153) &  0.6095 (0.0173) &  0.6165 (0.0255) &  0.8573 (0.0296) \\
& Ethnicity & & \\
& \quad Non-Hispanic & 0.5879 (0.0139) & 0.6299 (0.0100) & 0.6312 (0.0103) & 0.7291 (0.0420) \\
& \quad Hispanic & 0.5847 (0.0122) & 0.6300 (0.0063) & 0.6287 (0.0087) & 0.8230 (0.0227) \\
& Age & & \\
& \quad 18-24 & 0.5554 (0.0351) & 0.5960 (0.0178) & 0.6406 (0.0360) & 0.7462 (0.0415) \\
& \quad 25-54 & 0.5569 (0.0372) & 0.6014 (0.0187) & 0.6430 (0.0341) & 0.7334 (0.0439) \\
& \quad $>$ 55 & 0.5544 (0.0397) & 0.6011 (0.0205) & 0.6422 (0.0329) & 0.7824 (0.0389) \\

\bottomrule
\end{tabular}
\caption{All TPR results for predicting premature treatment discontinuation.}
\label{tab:all_TPR_dropout}
\end{table}

\begin{table}[htb]
\centering
\small
\begin{tabular}{@{}llcccc@{}}
\toprule
& & LR & RF & GBDT & MLP \\

\midrule
\multirow{10}{*}{Baseline} 
& Race & & \\
& \quad White & 0.1107 (0.0021) & 0.1114 (0.0016) & 0.1155 (0.0020) & 0.1288 (0.0397) \\
& \quad Black & 0.2459 (0.0050) & 0.2495 (0.0053) & 0.2656 (0.0068) & 0.2701 (0.0465) \\
& Ethnicity & & \\
& \quad Non-Hispanic & 0.1233 (0.0021) & 0.1264 (0.0016) & 0.1319 (0.0018) & 0.1471 (0.0401) \\
& \quad Hispanic & 0.2264 (0.0092) & 0.2023 (0.0102) & 0.2124 (0.0139) & 0.2422 (0.0505) \\
& Age & & \\
& \quad 18-24 & 0.1042 (0.0020) &  0.1084 (0.0022) &  0.1129 (0.0026) &  0.1320 (0.0403) \\
& \quad 25-54 & 0.1042 (0.0020) &  0.1084 (0.0022) &  0.1129 (0.0026) &  0.1320 (0.0403) \\
& \quad $>$ 55 & 0.4901 (0.0096) &  0.4131 (0.0110) &  0.4414 (0.0073) &  0.4584 (0.0693) \\

\midrule
\multirow{10}{*}{Reweighing} 
& Race & & \\
& \quad White & 0.1260 (0.0023) &  0.1227 (0.0013) &  0.1297 (0.0016) &  0.1543 (0.0322) \\
& \quad Black & 0.1403 (0.0029) &  0.1798 (0.0058) &  0.1795 (0.0075) &  0.2042 (0.0401) \\
& Ethnicity & & \\
& \quad Non-Hispanic & 0.1296 (0.0022) & 0.1286 (0.0017) & 0.1367 (0.0016) & 0.1612 (0.0310) \\
& \quad Hispanic & 0.1342 (0.0035) & 0.1547 (0.0058) & 0.1420 (0.0054) & 0.1561 (0.0375) \\
& Age & & \\
& \quad 18-24 & 0.1093 (0.0041) &  0.1041 (0.0043) &  0.1169 (0.0054) &  0.1430 (0.0294) \\
& \quad 25-54 & 0.1242 (0.0023) &  0.1282 (0.0021) &  0.1325 (0.0020) &  0.1607 (0.0265) \\
& \quad $>$ 55 & 0.1032 (0.0056) &  0.1394 (0.0054) &  0.1299 (0.0061) &  0.1613 (0.0479) \\

\midrule
\multirow{10}{*}{EGR} 
& Race & & \\
& \quad White & 0.1265 (0.0022) &  0.1329 (0.0040) &  0.1358 (0.0013) & -- \\
& \quad Black & 0.1386 (0.0038) &  0.1359 (0.0050) &  0.1435 (0.0047) & -- \\
& Ethnicity & & \\
& \quad Non-Hispanic & 0.1668 (0.0059) & 0.1343 (0.0053) & 0.1367 (0.0018) & -- \\
& \quad Hispanic & 0.1731 (0.0087) & 0.1451 (0.0086) & 0.1487 (0.0091) & -- \\
& Age & & \\
& \quad 18-24 & 0.1260 (0.0433) &  0.1374 (0.0080) &  0.1289 (0.0054) &  -- \\
& \quad 25-54 & 0.1083 (0.0376) &  0.1358 (0.0081) &  0.1291 (0.0022) &  -- \\
& \quad $>$ 55 & 0.1262 (0.0420) &  0.1620 (0.0102) &  0.1539 (0.0058) &  -- \\

\midrule
\multirow{10}{*}{Threshold} 
& Race & & \\
& \quad White & 0.3893 (0.0215) &  0.4040 (0.0112) &  0.3714 (0.0104) &  0.1377 (0.0261) \\
& \quad Black & 0.3925 (0.0239) &  0.4169 (0.0110) &  0.3775 (0.0104) &  0.2799 (0.0416) \\
& Ethnicity & & \\
& \quad Non-Hispanic & 0.4394 (0.0152) & 0.4071 (0.0095) & 0.3749 (0.0160) & 0.1517 (0.0275) \\
& \quad Hispanic & 0.4349 (0.0118) & 0.4285 (0.0125) & 0.3743 (0.0170) & 0.2339 (0.0417) \\
& Age & & \\
& \quad 18-24 & 0.5017 (0.0299) &  0.4006 (0.0201) &  0.4111 (0.0385) &  0.0804 (0.0196) \\
& \quad 25-54 & 0.4982 (0.0269) &  0.3957 (0.0177) &  0.4096 (0.0330) &  0.1401 (0.0274) \\
& \quad $>$ 55 & 0.5019 (0.0324) &  0.4349 (0.0184) &  0.4251 (0.0318) &  0.4469 (0.0575) \\

\bottomrule
\end{tabular}
\caption{All FPR results for predicting treatment retention $>$ 180 days.}
\label{tab:all_FPR_los}
\end{table}

\begin{table}[htb]
\centering
\small
\begin{tabular}{@{}llcccc@{}}
\toprule
& & LR & RF & GBDT & MLP \\

\midrule
\multirow{10}{*}{Baseline} 
& Race & & \\
& \quad White & 0.2480 (0.0022) & 0.2771 (0.0028) & 0.2940 (0.0039) & 0.2856 (0.0646) \\
& \quad Black & 0.4561 (0.0075) & 0.5026 (0.0041) & 0.5294 (0.0071) & 0.4991 (0.0623) \\
& Ethnicity & & \\
& \quad Non-Hispanic & 0.2727 (0.0023) & 0.3073 (0.0023) & 0.3262 (0.0032) & 0.3204 (0.0626) \\
& \quad Hispanic & 0.4165 (0.0108) & 0.4207 (0.0109) & 0.4363 (0.0168) & 0.4430 (0.0630) \\
& Age & & \\
& \quad 18-24 & 0.1110 (0.0057) & 0.2111 (0.0084) & 0.2226 (0.0093) & 0.1779 (0.0561) \\
& \quad 25-54 & 0.2356 (0.0294) & 0.2717 (0.0033) & 0.2885 (0.0047) & 0.2891 (0.0645) \\
& \quad $>$ 55 & 0.6729 (0.0072) & 0.6389 (0.0109) & 0.6723 (0.0092) & 0.6622 (0.0701) \\

\midrule
\multirow{10}{*}{Reweighing} 
& Race & & \\
& \quad White & 0.2734 (0.0020) & 0.2964 (0.0022) & 0.3176 (0.0027) & 0.3417 (0.0472) \\
& \quad Black & 0.2881 (0.0052) & 0.3951 (0.0058) & 0.3968 (0.0070) & 0.4248 (0.0621) \\
& Ethnicity & & \\
& \quad Non-Hispanic & 0.2829 (0.0022) & 0.3111 (0.0025) & 0.3338 (0.0030) & 0.3585 (0.0478) \\
& \quad Hispanic & 0.2745 (0.0031) & 0.3491 (0.0091) & 0.3307 (0.0102) & 0.3371 (0.0584) \\
& Age & & \\
& \quad 18-24 & 0.2381 (0.0064) & 0.2558 (0.0063) & 0.2859 (0.0074) & 0.3169 (0.0433) \\
& \quad 25-54 & 0.2691 (0.0021) & 0.3055 (0.0029) & 0.3208 (0.0029) & 0.3523 (0.0404) \\
& \quad $>$ 55 & 0.1837 (0.0071) & 0.2927 (0.0050) & 0.2774 (0.0099) & 0.3130 (0.0767) \\

\midrule
\multirow{10}{*}{EGR} 
& Race & & \\
& \quad White & 0.2741 (0.0022) & 0.3129 (0.0058) & 0.3272 (0.0025) & -- \\
& \quad Black & 0.2848 (0.0074) & 0.3189 (0.0081) & 0.3347 (0.0045) & -- \\
& Ethnicity & & \\
& \quad Non-Hispanic & 0.3290 (0.0094) & 0.3205 (0.0081) & 0.3338 (0.0032) & -- \\
& \quad Hispanic & 0.3185 (0.0140) & 0.3208 (0.0116) & 0.3385 (0.0101) & -- \\
& Age & & \\
& \quad 18-24 & 0.2002 (0.0744) & 0.2982 (0.0128) & 0.3038 (0.0086) &  -- \\
& \quad 25-54 & 0.2087 (0.0744) & 0.3161 (0.0113) & 0.3150 (0.0030) &  -- \\
& \quad $>$ 55 & 0.1891 (0.0724) & 0.3233 (0.0127) & 0.3122 (0.0065) &  -- \\

\midrule
\multirow{10}{*}{Threshold} 
& Race & & \\
& \quad White & 0.6052 (0.0210) & 0.6530 (0.0111) & 0.6232 (0.0100) & 0.3152 (0.0420) \\
& \quad Black & 0.6085 (0.0222) & 0.6616 (0.0098) & 0.6278 (0.0120) & 0.5340 (0.0552) \\
& Ethnicity & & \\
& \quad Non-Hispanic & 0.6509 (0.0115) & 0.6622 (0.0097) & 0.6337 (0.0147) & 0.3442 (0.0424) \\
& \quad Hispanic & 0.6513 (0.0192) & 0.6494 (0.0091) & 0.6312 (0.0216) & 0.4508 (0.0564) \\
& Age & & \\
& \quad 18-24 & 0.6807 (0.0263) & 0.6395 (0.0220) & 0.6513 (0.0351) & 0.2179 (0.0356) \\
& \quad 25-54 & 0.6814 (0.0269) & 0.6437 (0.0181) & 0.6567 (0.0323) & 0.3191 (0.0422) \\
& \quad $>$ 55 & 0.6820 (0.0303) & 0.6403 (0.0200) & 0.6524 (0.0360) & 0.6629 (0.0631) \\

\bottomrule
\end{tabular}
\caption{All TPR results for predicting treatment retention $>$ 180 days.}
\label{tab:all_TPR_los}
\end{table}

\subsection{Results for 2020--2023 TEDS-D Data} \label{results_20_23}
\begin{figure}[tbp]
\centering
\includegraphics[width=0.85\textwidth]{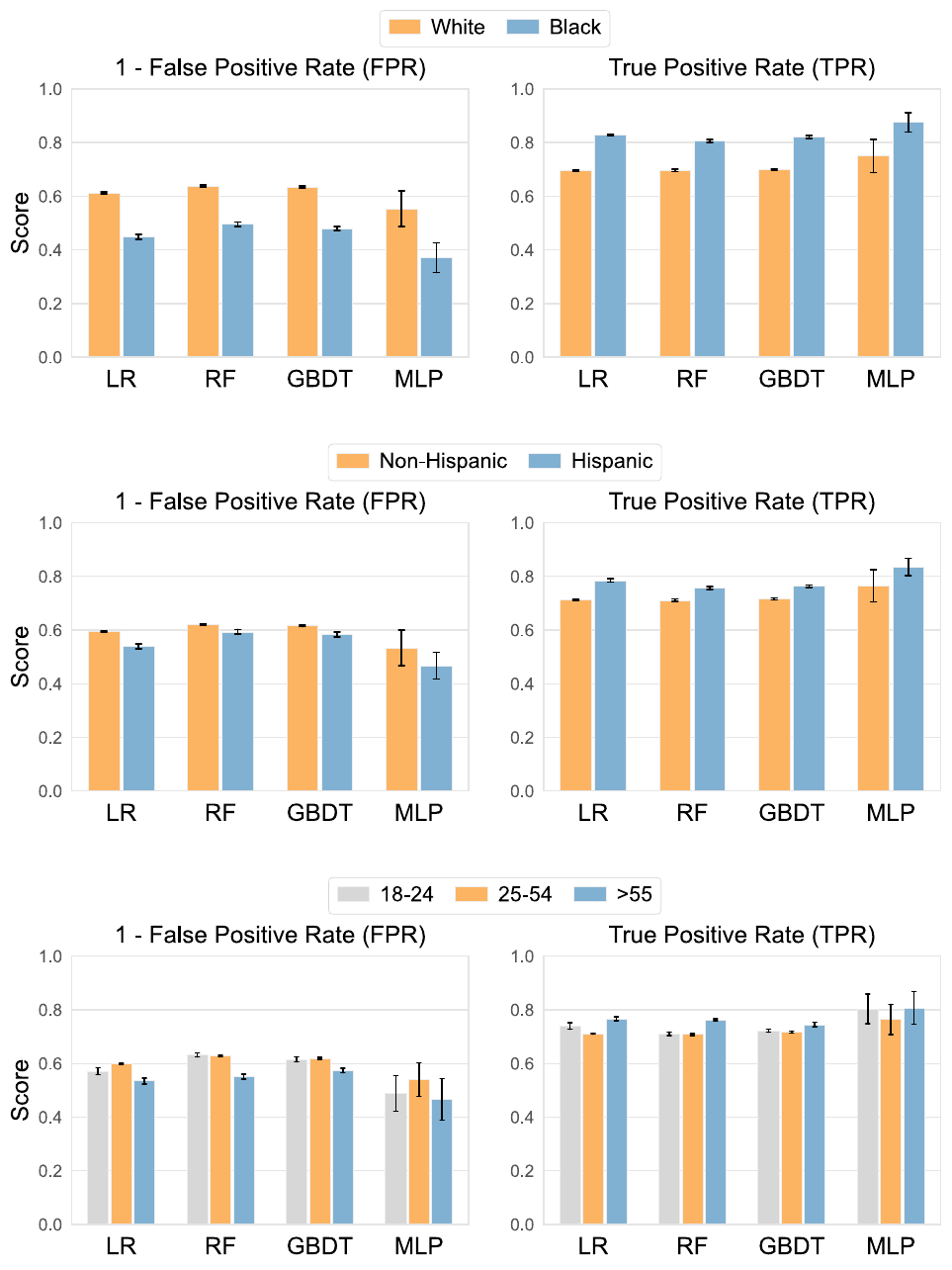}
\caption{FPRs and TPRs across different subgroups are evaluated for four ML models (LR, RF, GBDT, MLP) for predicting premature treatment discontinuation with 2020-2023 data. The subfigures presents results stratified by (a) race, (b) ethnicity, and (c) age. Bars represent the mean values over 10 runs, while error bars indicate the standard deviation computed from these runs. Higher values indicate better results. }
\label{fig:dropout_baseline_2023}
\end{figure}

\begin{figure}[tbp]
\centering
\includegraphics[width=0.85\textwidth]{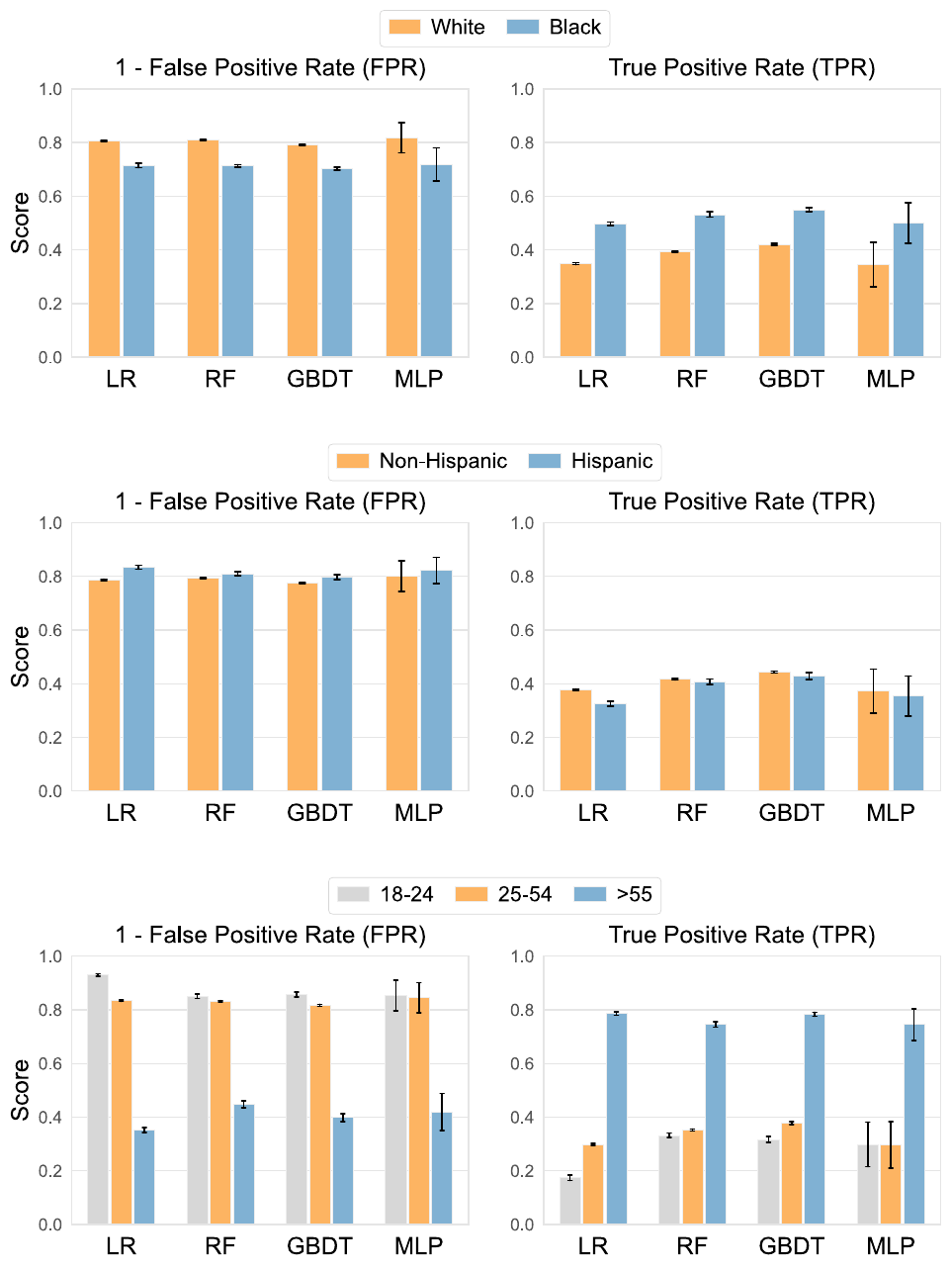}
\caption{FPRs and TPRs across different subgroups are evaluated for four ML models (LR, RF, GBDT, MLP) for predicting treatment retention $>$ 180 days with 2020-2023 data. The subfigures presents results stratified by (a) race, (b) ethnicity, and (c) age. Bars represent the mean values over 10 runs, while error bars indicate the standard deviation computed from these runs. Higher values indicate better results. }
\label{fig:los_baseline_2023}
\end{figure}

To examine whether the findings remained consistent in more recent years and to evaluate the potential impact of changes in MOUD treatment following the COVID-19 pandemic, we conducted additional experiments using the 2020--2023 TEDS-D data. The same cohort selection criteria, data preprocessing procedures, model development pipeline, and evaluation metrics were applied to the 2020--2023 cohort. 
We evaluated the predictive performance and fairness of the four models for predicting premature treatment discontinuation and treatment retention exceeding 180 days. The results are summarized in Table~\ref{tab:overall_performance_2023}, Figures~\ref{fig:dropout_baseline_2023}, and Figure~\ref{fig:los_baseline_2023}. Overall, the models achieved slightly better predictive performance than those obtained using the 2015--2019 cohort. In addition, the fairness evaluation based on the Equalized Odds Difference (EOD) exhibited patterns similar to those observed in the primary analysis, with slightly smaller disparities across demographic subgroups. Overall, the results demonstrate that the proposed modeling framework produces consistent predictive performance and fairness characteristics when applied to more recent TEDS-D data.

\begin{table}[tbp]
\caption{Overall predictive performance, measured by area under the receiver operating characteristic curve (AUROC), as well as Equalized Odds Difference (EOD) across subgroups for ML models predicting premature treatment discontinuation and treatment retention exceeding 180 days using 2020-2023 TEDS-D data.}
\label{tab:overall_performance_2023}
\begin{tabular*}{0.85\textwidth}{@{\extracolsep\fill}llcc}
\toprule 
& Model & Premature Discontinuation & Retention $>$ 180 days \\
\midrule
\multirow{4}{*}{AUROC}
& LR & 0.7153 (0.0009) & 0.6427 (0.0015) \\
& RF & 0.7367 (0.0011) & 0.6770 (0.0017) \\
& GBDT & 0.7378 (0.0012) & 0.6781 (0.0018) \\
& MLP & 0.7173 (0.0022) & 0.6555 (0.0039) \\
\midrule
\multirow{4}{*}{EOD - Race}
& LR & 0.1633 (0.0110) & 0.1486 (0.0081) \\
& RF & 0.1425 (0.0094) & 0.1390 (0.0100) \\
& GBDT & 0.1538 (0.0090) & 0.1290 (0.0077) \\
& MLP & 0.1824 (0.0273) & 0.1549 (0.0382) \\
\midrule
\multirow{4}{*}{EOD - Ethnicity}
& LR & 0.0722 (0.0068) & 0.0525 (0.0090) \\
& RF & 0.0446 (0.0080) & 0.0154 (0.0081) \\
& GBDT & 0.0467 (0.0080) & 0.0218 (0.0087) \\
& MLP & 0.0702 (0.0356) & 0.0274 (0.0233) \\
\midrule
\multirow{4}{*}{EOD - Age}
& LR & 0.0637 (0.0112) & 0.6118 (0.0092) \\
& RF & 0.0829 (0.0134) & 0.4134 (0.0168) \\
& GBDT & 0.0471 (0.0101) & 0.4661 (0.0123) \\
& MLP & 0.0779 (0.0191) & 0.4601 (0.0367) \\
\bottomrule
\end{tabular*}
\end{table}

\end{appendices}


\bibliography{sn-bibliography}

\end{document}